\documentclass[11pt]{article}
\usepackage[body={7in, 9in},left=1in,right=1in]{geometry}
\usepackage[small]{titlesec} %

\usepackage[arxiv]{optional} 

\usepackage{enumerate}
\usepackage{latexsym}
\usepackage{graphicx}
\usepackage{multicol,multirow}
\usepackage{amsmath,amssymb,amsfonts}
\usepackage{mathrsfs}
\usepackage{amsthm}
\usepackage{appendix}

\usepackage[numbers]{natbib}

\usepackage[T1]{fontenc}
\usepackage{times}
\usepackage{textcomp}%
\usepackage{url}
\usepackage{comment}
\usepackage{xspace}
\usepackage{mathtools}
\usepackage{afterpage}
\usepackage{capt-of}
\usepackage{subcaption}
\usepackage[colorlinks,citecolor=blue,urlcolor=blue,linkcolor=blue,pdfborder={0 0 0}]{hyperref}
\usepackage[capitalize]{cleveref}

\crefname{figure}{Figure}{Figures}
\usepackage{booktabs}
\usepackage[dvipsnames,table,xcdraw]{xcolor}
\usepackage{gensymb}
\usepackage{algorithm}
\usepackage[noend]{algorithmic} %
\usepackage{soul}
\usepackage{tabularx}
\usepackage[export]{adjustbox}
\usepackage{enumitem}

\usepackage[resetlabels]{multibib}
\newcites{SM}{Supplementary References}

\graphicspath{{figures/}}

\def\R{\mathbb{R}}
\newcommand{\US}{U.S.\xspace}

\newcommand{\pp}{\texttt{++}}
\newcommand{\cfs}{CFSv2\xspace}

\newcommand{\climpp}{Climatology\pp\xspace}

\newcommand{\perpp}{Persistence\pp\xspace}
\newcommand{\ecmwf}{ECMWF\xspace}

\newcommand{\aecmwf}{ABC-ECMWF\xspace}
\newcommand{\nwp}{dynamical\xspace}
\newcommand{\nwppp}{Dynamical\pp\xspace}

\newcommand{\cmss}[1]{{\fontfamily{cmss}\selectfont #1}}
\newcommand{\dataset}{\cmss{SubseasonalClimateUSA}\xspace} %
\newcommand{\toolkitpackage}{\texttt{subseasonal\_toolkit}\xspace} %
\newcommand{\datapackage}{\texttt{subseasonal\_data}\xspace} %

\def\rmse{\mathrm{RMSE}}
\def\mse{\mathrm{MSE}}
\def\skill{\mathrm{skill}}
\newcommand{\gt}{\mathbf{y}}
\newcommand{\predgt}{\hat{\mathbf{y}}}

\newcommand{\trainset}{\mc{T}}

\newcommand{\algorithmicinput}{\textbf{input}}
\newcommand{\INPUT}{\item[\algorithmicinput]}
\newcommand{\algorithmicoutput}{\textbf{output}}
\newcommand{\OUTPUT}{\item[\algorithmicoutput]}

\newcommand{\fcst}{\mbf{f}}
\newcommand{\leads}{\mathcal{L}}
\newcommand{\lstar}{l^\star}
\newcommand{\climvec}{\mbf{c}}

\newcommand{\INIT}{\ENSURE}
\newcommand{\loss}{\textup{loss}}
\newcommand{\dstar}{d^{\star}}
\newcommand{\tstar}{t^{\star}}

\DeclarePairedDelimiter\floor{\lfloor}{\rfloor}
\def\balign#1\ealign{\begin{align}#1\end{align}}
\def\baligns#1\ealigns{\begin{align*}#1\end{align*}}
\def\balignat#1\ealign{\begin{alignat}#1\end{alignat}}
\def\balignats#1\ealigns{\begin{alignat*}#1\end{alignat*}}
\def\bitemize#1\eitemize{\begin{itemize}#1\end{itemize}}
\def\benumerate#1\eenumerate{\begin{enumerate}#1\end{enumerate}}

\newenvironment{talign*}
 {\csname align*\endcsname}
 {\endalign}
\newenvironment{talign}
 {\csname align\endcsname}
 {\endalign}

\def\balignst#1\ealignst{\begin{talign*}#1\end{talign*}}
\def\balignt#1\ealignt{\begin{talign}#1\end{talign}}
\let\originalleft\left
\let\originalright\right
\renewcommand{\left}{\mathopen{}\mathclose\bgroup\originalleft}
\renewcommand{\right}{\aftergroup\egroup\originalright}

\newcommand{\Nino}{Ni\~no\xspace}

\def\tinycitep*#1{{\tiny\citep*{#1}}}
\def\tinycitealt*#1{{\tiny\citealt*{#1}}}
\def\tinycite*#1{{\tiny\cite*{#1}}}
\def\smallcitep*#1{{\scriptsize\citep*{#1}}}
\def\smallcitealt*#1{{\scriptsize\citealt*{#1}}}
\def\smallcite*#1{{\scriptsize\cite*{#1}}}

\def\mbi#1{\boldsymbol{#1}} %
\def\mbf#1{\mathbf{#1}}
\def\mbb#1{\mathbb{#1}}
\def\mc#1{\mathcal{#1}}

\def\tbf#1{\textbf{#1}}

\def\textsum{{\textstyle\sum}} %
\def\R{\mathbb{R}}
\def\<{\left\langle} %
\def\>{\right\rangle}

\newcommand{\textfrac}[2]{{\textstyle\frac{#1}{#2}}}

\def\indic#1{\mbb{I}\left[{#1}\right]} %
\providecommand{\argmin}{\mathop\mathrm{arg min}}

\ifdefined\nonewproofenvironments\else
\ifdefined\ispres\else

\newenvironment{proof-sketch}{\noindent\textbf{Proof Sketch}
  \hspace*{1em}}{\qed\bigskip\\}
\newenvironment{proof-idea}{\noindent\textbf{Proof Idea}
  \hspace*{1em}}{\qed\bigskip\\}
\newenvironment{proof-of-lemma}[1][{}]{\noindent\textbf{Proof of Lemma {#1}}
  \hspace*{1em}}{\qed\\}
\newenvironment{proof-of-theorem}[1][{}]{\noindent\textbf{Proof of Theorem {#1}}
  \hspace*{1em}}{\qed\\}
\newenvironment{proof-attempt}{\noindent\textbf{Proof Attempt}
  \hspace*{1em}}{\qed\bigskip\\}

\fi

\fi
\newcommand\blfootnote[1]{%
  \begingroup
  \renewcommand\thefootnote{}\footnote{#1}%
  \addtocounter{footnote}{-1}%
  \endgroup
}

\begin{document}
\opt{submit}{
\renewcommand{\includegraphics}[2][]{}
}

\title{%
Adaptive Bias Correction for Improved Subseasonal Forecasting\\ \ \\
}

\begin{center}
\Large
    \textbf{Adaptive Bias Correction for Improved Subseasonal Forecasting}
\end{center}

\vspace{.5\baselineskip}
\begin{center}
\textbf{Soukayna Mouatadid$^1$,
Paulo Orenstein$^2$,
Genevieve Flaspohler$^{3,4,5}$,
Judah Cohen$^{6,7}$,
Miruna Oprescu$^8$,
Ernest Fraenkel$^9$,
Lester Mackey$^{10}$}
\end{center}

\begin{center}
\footnotesize
$^1${Department of Computer Science, University of Toronto, Toronto, ON, Canada}\\
$^2${Instituto de Matem\'atica Pura e Aplicada, Rio de Janeiro, Brazil}\\
$^3${\textit{n}Line Inc., Berkeley, CA, USA}\\
$^4${Department of Electrical Engineering and Computer Science, Massachusetts Institute of Technology, Cambridge, MA, USA}\\
$^5${Department of Applied Ocean Science and Engineering, Woods Hole Oceanographic Institution, Falmouth, MA, USA}\\
$^6${Atmospheric and Environmental Research, Lexington, MA, USA}\\
$^7${Department of Civil and Environmental Engineering, Massachusetts Institute of Technology, Cambridge, MA, USA}\\
$^8${Department of Computer Science, Cornell University, Ithaca, NY, USA}\\
$^9${Department of Biological Engineering, Massachusetts Institute of Technology, Cambridge, MA, USA}\\
$^{10}${Microsoft Research New England, Cambridge, MA, USA}\blfootnote{Corresponding authors: Soukayna Mouatadid, \href{mailto:soukayna@cs.toronto.edu}{soukayna@cs.toronto.edu}; Lester Mackey, \href{mailto:lmackey@microsoft.com}{lmackey@microsoft.com}}
\end{center}
\vspace{0\baselineskip}

\newpage
\begin{abstract}
Subseasonal forecasting---predicting temperature and precipitation 2 to 6 weeks ahead---is critical for effective water allocation, wildfire management, and drought and flood mitigation. 
Recent international research efforts have advanced the subseasonal capabilities of operational dynamical models, yet temperature and precipitation prediction skills remain poor, 
partly due to stubborn errors in representing atmospheric dynamics and physics inside dynamical models.
Here, to counter these errors, we introduce an \textit{adaptive bias correction} (ABC) method that combines state-of-the-art dynamical forecasts with observations using machine learning.
We show that, when applied to the leading subseasonal model from the European Centre for Medium-Range Weather Forecasts (ECMWF), ABC improves temperature forecasting skill by 60-90\% (over baseline skills of 0.18-0.25) and precipitation forecasting skill by 40-69\% (over baseline skills of 0.11-0.15) in the contiguous U.S. 
We couple these performance improvements with a practical workflow 
to explain ABC skill gains and identify higher-skill windows of opportunity based on specific climate conditions.
\end{abstract}

\section{Introduction} \label{sec:introduction}
	Improving our ability to forecast both weather and climate is of interest to many sectors of the economy and government agencies, from the local to the national level. Weather forecasts 0 to 10 days ahead and climate forecasts seasons to decades ahead are currently used operationally in decision-making, and the accuracy and reliability of these forecasts has improved consistently in recent decades \citep{troccoli2010a}. However, many critical applications---including water allocation, wildfire management, and drought and flood mitigation---require \emph{subseasonal forecasts}, with lead times beyond 10 days and
up to a season \citep{merryfield2020current, white2017}.  Given the changing nature of the climate and the increasing frequency of extreme weather events, there is a social and scientific consensus regarding the importance and urgency of providing reliable subseasonal forecasts~\citep{board2016next,mariotti2020windows}.

Subseasonal forecasting lies in a challenging intermediate domain between shorter-term weather forecasting (an initial value problem) and longer-term climate forecasting (a boundary value problem).
Skillful subseasonal forecasting requires capturing the complex dependence between local weather conditions, typically described by numerical weather models, and global climate variables, usually part of long-range seasonal forecasts \citep{merryfield2020current}. The intertwined dynamics of initial-value prediction problems and boundary forcing phenomena led subseasonal forecasting to long be considered a 
predictability desert~\citep{vitart2012subseasonal}, more difficult than either short-term weather forecasting or long-term climate prediction. Recent studies, however, have highlighted important sources of predictability on subseasonal timescales, including oscillatory modes such as El \Nino-Southern Oscillation and the Madden-Julian oscillation, large-scale anomalies in, e.g., soil moisture or sea ice, and external forcing \citep{board2016next, l2021sources}. 
These predictability sources are imperfectly understood and imperfectly represented in weather and climate models~\citep{reyniers2008quantitative} and hence represent an opportunity for more skillful subseasonal forecasting.
   	
    The challenges of subseasonal forecasting are particularly apparent for precipitation forecasts~\citep{vitart2017subseasonal}. Precipitation is governed by both macro-scale dynamical processes of the atmosphere and complex microphysical processes, some of which are still not fully understood~\citep{reyniers2008quantitative}. In addition, precipitation is oftentimes a very local phenomenon, working over a much finer scale than the resolution employed by subseasonal dynamical models.
    This scale incompatibility, in concert with  suboptimal process representation and incomplete process understanding, results in dynamical models falling short of predictability limits for forecasting precipitation~\citep{reyniers2008quantitative,board2016next}. 
    Generating precipitation forecasts for longer seasonal horizons is an even more daunting task. In this case, dynamical models show considerable biases in precipitation and wind fields. These biases arise from the parameterization of key physical processes associated with deep convective cloud systems~\citep{chantry2021opportunities} and, when combined with chaotic dynamics and imperfectly represented sources of predictability,  translate into rapidly decreasing skill for precipitation forecasts.

    Bridging the gap between short-term and seasonal forecasting has been the focus of several recent large-scale research efforts to advance the subseasonal capabilities of operational physics-based models~\citep{vitart2017subseasonal,pegion2019subseasonal,lang2020introduction}. However, despite these advances, dynamical models still suffer from persistent systematic errors, which limit the skill of temperature and precipitation forecasts for longer subseasonal lead times. 
    Low skill at these time horizons has a palpable practical impact on the utility of subseasonal forecasts for policy planners and stakeholders.

    To counter the observed systematic errors of physics-based models on the subseasonal timescale, there have been parallel efforts in recent years to demonstrate the value of machine learning and 
    deep learning methods for improved subseasonal forecasting accuracy \citep{li2016implications, cohen2018a, hwang2019improving, arcomano2020machine, he2021sub, yamagami2020subseasonal, wang2021improving, kim2021enhancing, kim2021deep, fan2021using, watson2021machine, weyn2021sub, srinivasan2021subseasonal}. 
    While these works demonstrate the promise of learned models for subseasonal forecasting, they also highlight the complementary strengths of physics- and learning-based approaches and the opportunity to combine those strengths to improve forecasting skill~\citep{hwang2019improving,kim2021enhancing,kim2021deep,fan2021using}. 
    
    To harness the complementary strengths of physics- and learning-based models, 
    we introduce a hybrid dynamical-learning framework  for improved subseasonal forecasting.
	In particular, we learn to adaptively correct the biases of dynamical models and 
	apply our \emph{adaptive bias correction} (ABC) 
	to improve the skill of subseasonal temperature and precipitation forecasts.
	ABC 
	is an ensemble of three low-cost, high-accuracy machine learning models introduced in this work:
	\nwppp, \climpp, and \perpp.
	Each model trains only on past temperature, precipitation, and forecast data and outputs corrections for future forecasts tailored to the site, target date, and dynamical model.  
    \nwppp and \climpp learn site- and date-specific offsets for dynamical and climatological forecasts by minimizing forecasting error over adaptively-selected training periods. 
    \perpp 
    additionally accounts for recent weather trends by combining lagged observations, dynamical  forecasts, and climatology to minimize historical forecasting error for each site.
    More details on each component model can be found in \cref{sec:methods}.
    Correction alone is no substitute for improved understanding and representation of predictability sources, and we therefore view ABC as a complement for improved dynamical modeling.
    Fortunately, 
    as an adaptive correction, 
    ABC automatically benefits from scientific improvements to its dynamical model inputs while learning to compensate for their residual systematic errors.
  	
    ABC can be applied operationally as a computationally inexpensive enhancement to any dynamical model forecast, 
	and we use this property to substantially reduce the forecasting errors of eight operational dynamical models, including the state-of-the-art \ecmwf model.
    ABC also improves upon the skill of classical and recently-developed bias corrections from the subseasonal forecasting literature including quantile mapping~\citep{panofsky1968some,monhart2018skill,baker2019developing,li2019evaluation}, locally estimated scatterplot smoothing  (LOESS)~\citep{cleveland1988locally,monhart2018skill}, and neural network~\citep{fan2021using} approaches.
	We couple these performance improvements with a practical workflow for explaining 
	ABC skill gains 
	using Cohort Shapley~\citep{mase2019explaining}
	and identifying higher-skill windows of opportunity~\citep{mariotti2020windows} based on relevant climate variables.
	To facilitate future deployment and development, we release our ABC model and workflow code through the \href{https://github.com/microsoft/subseasonal\_toolkit}{\toolkitpackage} Python package.
\section{Results} \label{sec:results}

    \subsection*{Improved precipitation and temperature prediction with adaptive bias correction}

    \begin{figure}[th!]
        \centering
        \includegraphics[width=\textwidth]{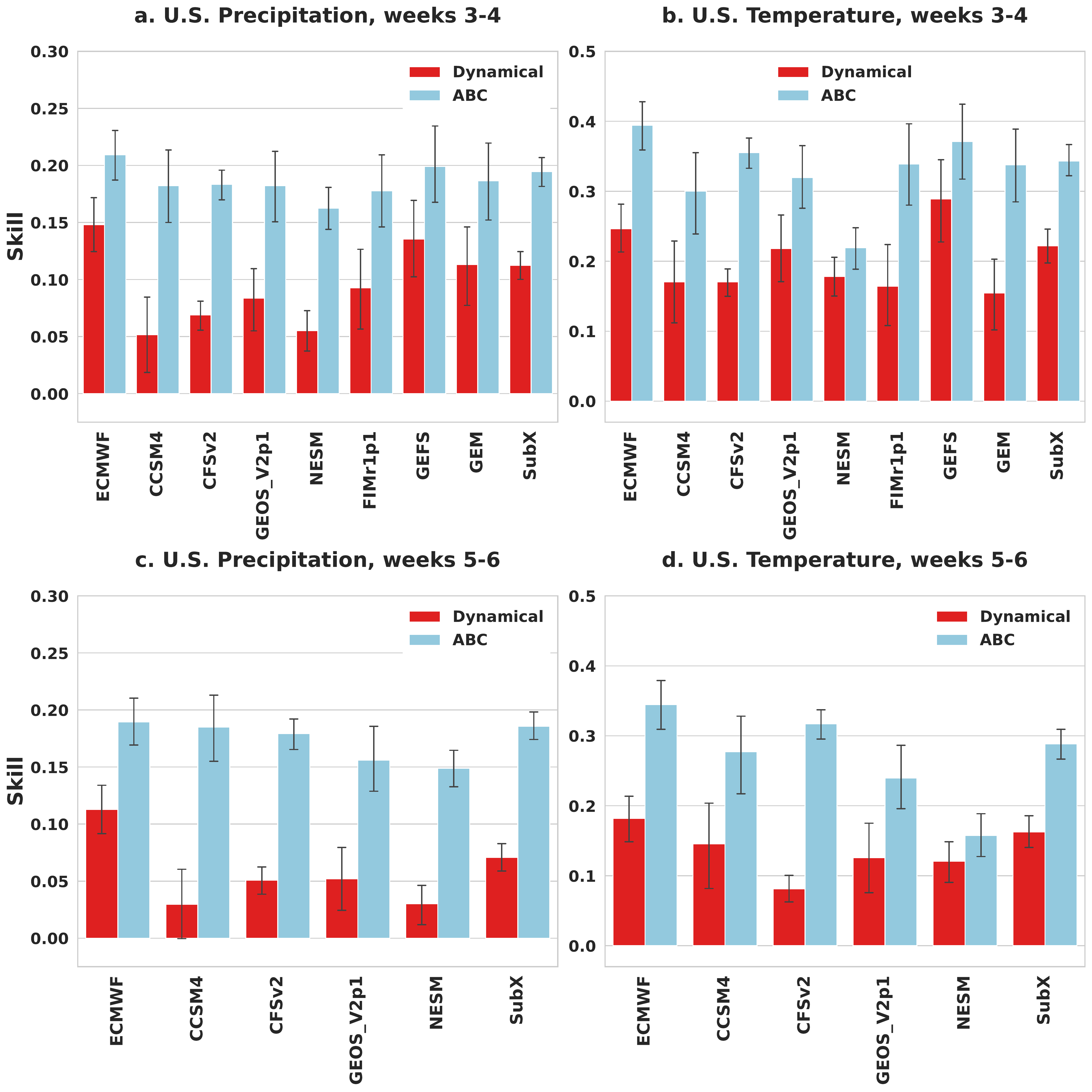}
        \caption{\textbf{Average forecast skill for dynamical models (red) and their adaptive bias correction (ABC) counterparts (blue).}
        Across the contiguous \US and the years 2018--2021, ABC provides a pronounced improvement in skill for each SubX or ECMWF dynamical model input 
        and each forecasting task (\textbf{a, b, c, d}). 
        The error bars display 95\% bootstrap confidence intervals. Models without forecast data for weeks 5-6 are omitted from the bottom panels.}
        \label{fig:barplot_subx_vs_abc}
    \end{figure}

    \begin{figure}[t!]
    \centering
    \includegraphics[width=\textwidth]{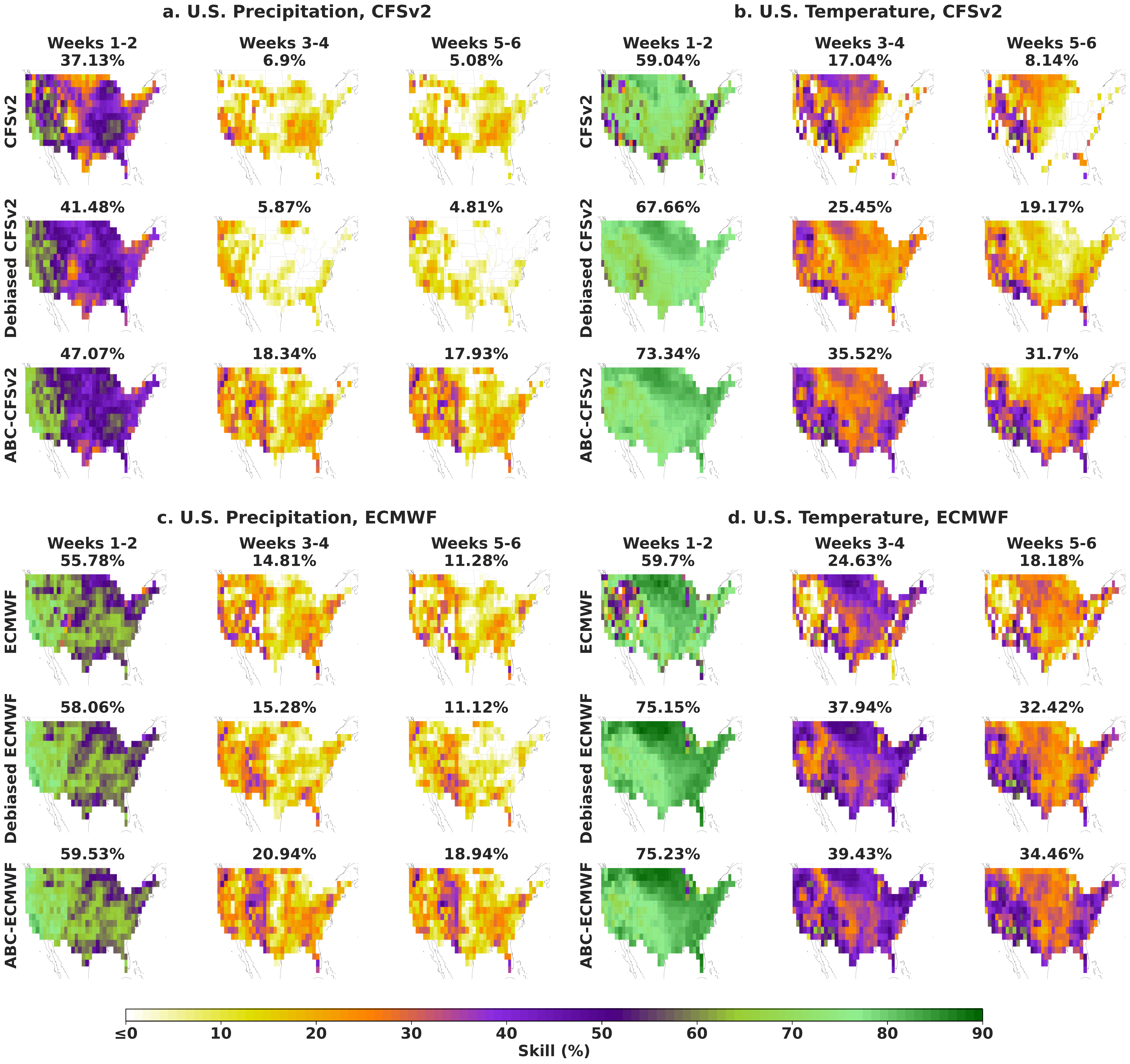}
    \caption{\textbf{Spatial skill distribution of dynamical models and their adaptive bias corrections.} 
    Across the contiguous \US and the years 2018--2021, dynamical model skill drops precipitously at subseasonal timescales (weeks 3-4 and 5-6), but 
    adaptive bias correction (ABC) attenuates the degradation, doubling or tripling the skill of \cfs (\textbf{a, b})  and boosting \ecmwf skill 40-90\%  over baseline skills of 0.11-0.25 (\textbf{c, d}).
    Taking the same raw model forecasts as input, ABC also provides consistent improvements over operational debiasing protocols, tripling the precipitation skill of debiased \cfs and improving that of debiased \ecmwf by 70\% (over a baseline skill of 0.11).
    The average temporal skill over all forecast dates is displayed above each map.}
    \label{fig:skill_spatial_distribution}
    \end{figure} 

        \begin{figure}[th!]
        \centering
        \includegraphics[width=\textwidth]{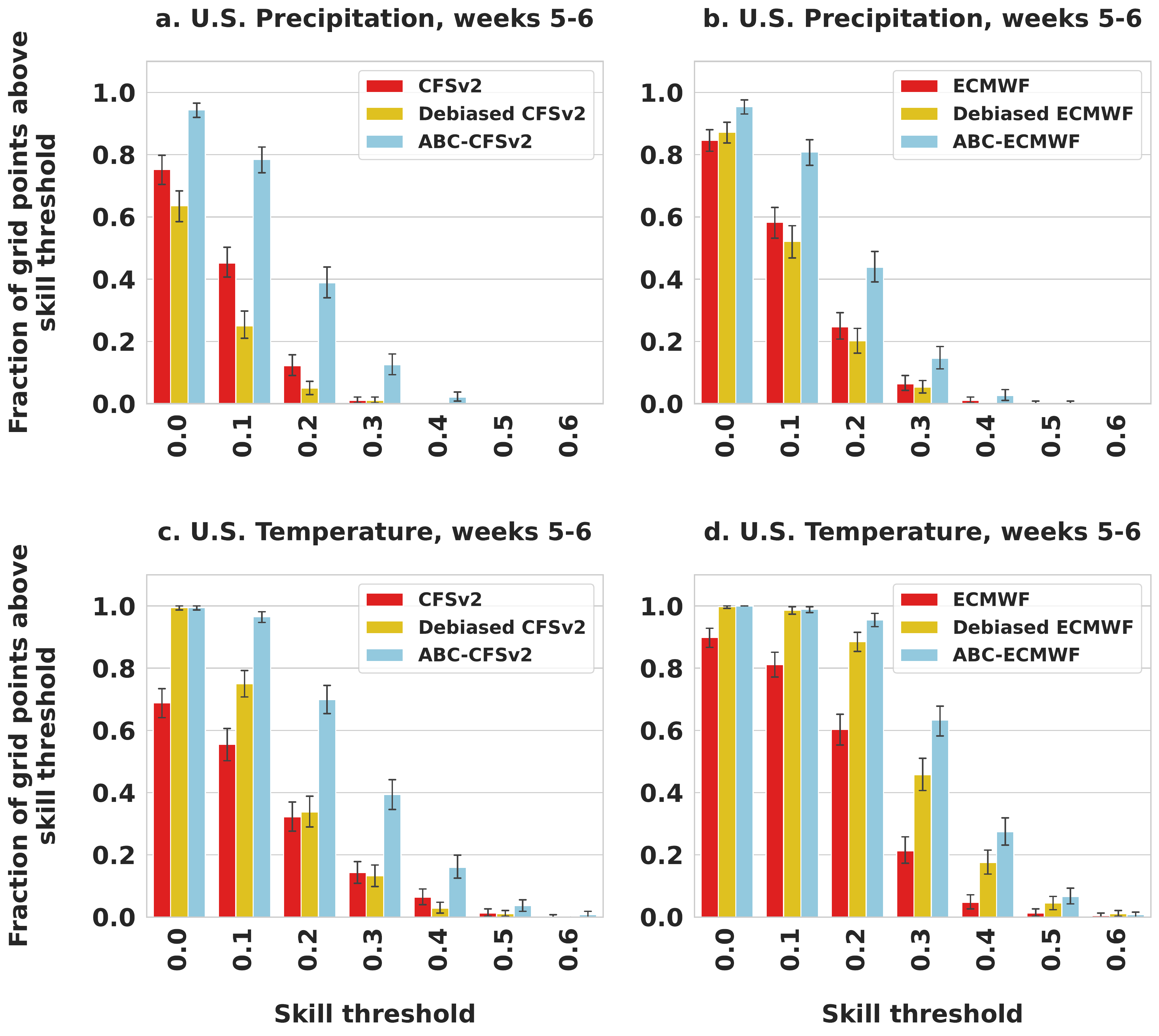}
    \caption{\textbf{Fraction of contiguous \US 
    with 2018--2021 spatial skill above a given threshold.}
    For each forecasting task and dynamical model input (\textbf{a, b, c, d}), 
    adaptive bias correction (ABC) consistently 
    expands the geographic range of 
    higher skill over raw and operationally-debiased dynamical models.
    The error bars display 95\% bootstrap confidence intervals. 
    }    
    \label{fig:barplot_fraction_above_treshold_skill_56w}
    \end{figure}
    
    \cref{fig:barplot_subx_vs_abc} highlights the advantage of ABC over raw dynamical models when forecasting accumulated precipitation and averaged temperature in the contiguous U.S. 
    Here, ABC is applied to the leading subseasonal model, ECMWF, to each of seven operational models participating in the Subseasonal Experiment \citep[SubX,][]{pegion2019subseasonal}, and to the mean of the SubX models. 
    Subseasonal forecasting skill, measured by uncentered anomaly correlation, is evaluated at two forecast horizons, weeks 3-4 and weeks 5-6, and averaged over all available forecast dates in the four-year span 2018--2021. 
    We find that, for each dynamical model input and forecasting task, ABC  leads to a pronounced improvement in skill.
    For example, when applied to the \US operational Climate Forecast System  Version 2 (\cfs),
    ABC improves temperature forecasting skill by 109-289\% (over baseline skills of 0.08-0.17) and precipitation skill by 165-253\% (over baseline skills of 0.05-0.07). When applied to the leading \ecmwf model, ABC improves temperature skill by 60-90\% (over baseline skills of 0.18-0.25) and precipitation  skill by 40-69\% (over baseline skills of 0.11-0.15).
    Moreover, for precipitation, even lower-skill models like CCSM4 have improved skill that is comparable to the best dynamical model after the application of ABC. 
    Overall---despite significant variability in dynamical model skill---ABC consistently reduces the systematic errors of its input model, bringing forecasts closer to observations for each target variable and time horizon.
    Similar forecast improvement is observed when stratifying skill by season (see \opt{submit}{Figure~S1}\opt{arxiv}{\cref{fig:barplot_subx_vs_abc_quarterly}}).
    
    In \opt{submit}{Figure~S2}\opt{arxiv}{\cref{fig:barplot_nn-a}}, we compare ABC with three additional subseasonal debiasing baselines (detailed in \cref{sec:methods}): quantile mapping \citep{panofsky1968some,monhart2018skill,baker2019developing,li2019evaluation}, LOESS debiasing  \citep{cleveland1988locally,monhart2018skill}, and a recently proposed neural network debiasing scheme trained jointly on temperature and precipitation inputs \citep[NN-A,][]{fan2021using}.  
    Given the same dynamical model inputs, ABC improves upon the skill of each baseline for each target variable and forecast horizon.
    
    We next examine the spatial distribution of skill for \cfs, \ecmwf, and their ABC-corrected counterparts at three forecast horizons in \cref{fig:skill_spatial_distribution}.
    At the shorter-term horizon of weeks 1-2, both \cfs and \ecmwf have reasonably high skill throughout the contiguous U.S. 
    However, skill drops precipitously for both models when moving to the subseasonal forecast horizons (weeks 3-4 and 5-6).
    This degradation is particularly striking for precipitation, where prediction skill drops to zero or to negative values in the central and northeastern parts of the U.S. 
    For temperature prediction, \cfs has a skill of zero across a broad region of the East,
    while \ecmwf produces isolated pockets of zero skill in the West.
    At these subseasonal timescales, ABC provides consistent improvements across the \US that either double or triple the mean skill of \cfs and increase the mean skill of \ecmwf by 40-90\% (over baseline skills of 0.11-0.25).
    Similar improvements are observed when ABC is applied to the SubX multimodel mean (see \opt{submit}{Figure~S3}\opt{arxiv}{\cref{fig:skill_spatial_distribution_subx_mean}}).
    In addition, common skill patterns across models are apparent that are consistent with higher precipitation predictability in the Western \US than in the Eastern \US and higher temperature predictability on the coasts than in the center of the country.

    Notably, ABC also improves over standard operational debiasing protocols (labeled \texttt{debiased \cfs} and \texttt{debiased \ecmwf} in \cref{fig:skill_spatial_distribution}), tripling the average precipitation skill of debiased \cfs and increasing that of debiased \ecmwf by 70\% (over a baseline skill of 0.11).
    As seen in \opt{submit}{Figure~S4}\opt{arxiv}{\cref{fig:skill_spatial_distribution_cfsv2}}, ABC additionally improves upon the quantile mapping, LOESS, and neural network debiasing baselines, doubling the ECMWF precipitation skill of the best-performing baseline and improving the ECMWF temperature skill by 37\% (over a baseline skill of 0.25).

    A practical implication of these improvements for downstream decision-makers is an expanded geographic range for 
    actionable skill, defined here as spatial skill above a given sufficiency threshold.
    For example, in \cref{fig:barplot_fraction_above_treshold_skill_56w}, we vary  
    the weeks 5-6 sufficiency threshold from $0$ to $0.6$ 
    and find that
    ABC consistently boosts 
    the number of locales with actionable skill 
    over both raw and operationally-debiased \cfs and ECMWF.
    We observe similar gains for weeks 3-4 in \opt{submit}{Figure~S5}\opt{arxiv}{\cref{fig:barplot_fraction_above_treshold_skill_34w}} and for ABC correction of the SubX multimodel mean in \opt{submit}{Figure~S6}\opt{arxiv}{\cref{fig:barplot_fraction_above_treshold_skill_subx_mean}}. 

    We emphasize that our results, like those of \cite{monhart2018skill,baker2019developing,li2019evaluation,fan2021using,kim2021deep}, focus on improved \emph{deterministic} forecasting: outputting a more accurate point estimate of a future weather variable.
    The complementary paradigm of probabilistic forecasting instead predicts the distribution of a weather variable, i.e., the probability that a variable will fall above or below any given threshold.
    Ideally, one would employ a tailored approach to probabilistic debiasing that directly optimizes a probabilistic skill metric to output a corrected distribution.  However, there is a simple, inexpensive way to convert the output of ABC into a probabilistic forecast.  Given any ensemble of  dynamical model forecasts (e.g., the control and perturbed forecasts routinely generated operationally), one can train ABC on the ensemble mean, apply the learned bias corrections to each ensemble member individually,  and use the empirical distribution of those bias corrected forecasts as the probabilistic forecasting estimate.  In \opt{submit}{Figures~S7, S8, S9, and S10}\opt{arxiv}{\cref{fig:barplot_bss_quaterly,,fig:barplot_crps_quaterly,,fig:barplot_bss_quaterly_loess,,fig:barplot_crps_quaterly_loess}}, we present two standard probabilistic skill metrics---the continuous ranked probability score (CRPS) 
 and the Brier skill score (BSS) for above normal observations~\citep{hersbach2000decomposition}, defined in \opt{submit}{Supplementary Methods}\opt{arxiv}{\nameref{sec:supp_methods}}---and observe that, for each target variable, forecast horizon, and season, ABC improves upon the BSS and CRPS of \ecmwf, LOESS debiasing, and quantile mapping debiasing.

\begin{figure}[!t]
    \centering
    \includegraphics[width=\textwidth]{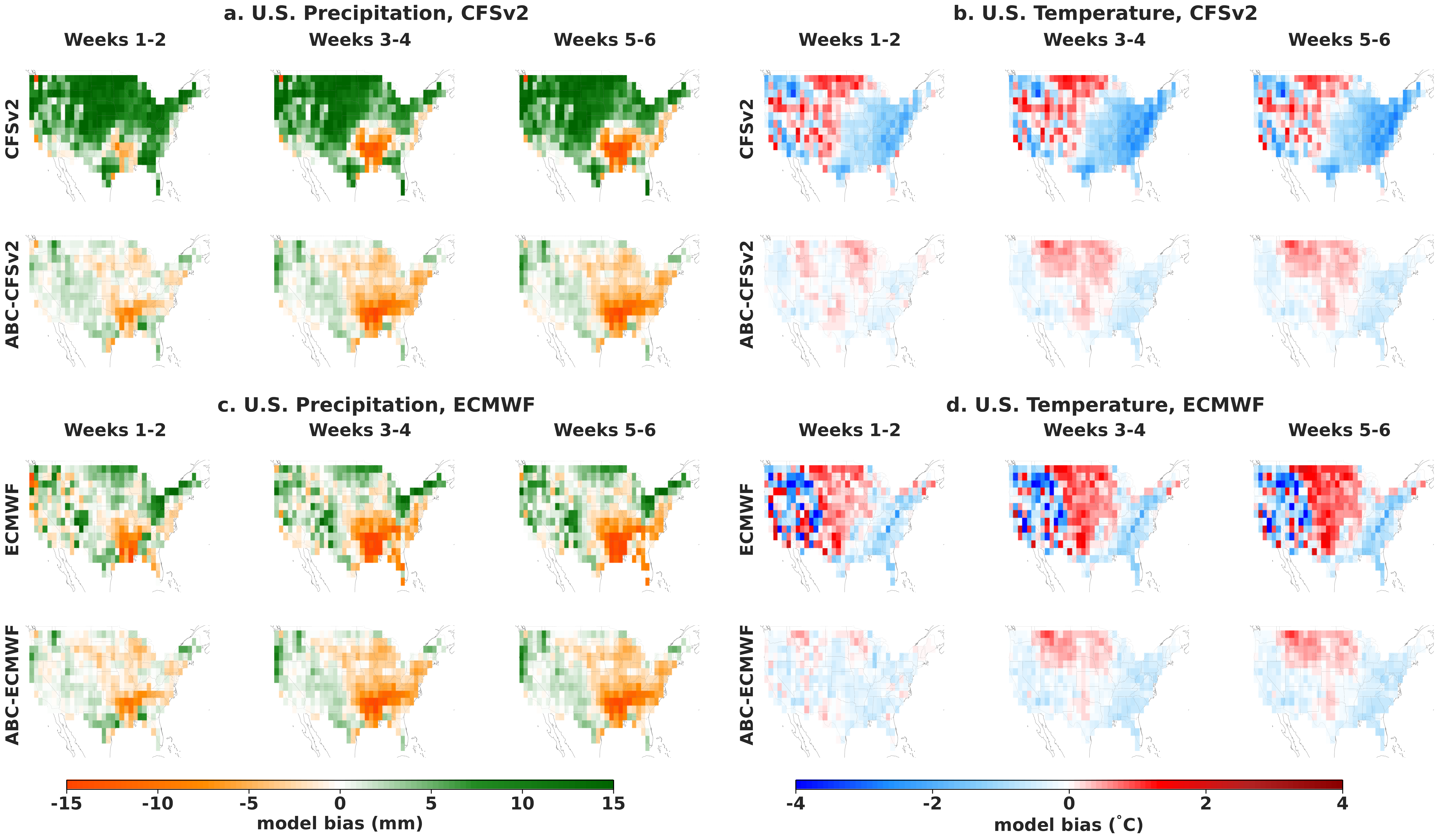}
    \caption{\textbf{Spatial distribution of model bias over the years 2018--2021.} Across the contiguous \US, adaptive bias correction (ABC) reduces the systematic model bias of its dynamical model input for both precipitation (\textbf{a, c}) and temperature (\textbf{b, d}).}
    \label{fig:bias_spatial_distribution}
    \end{figure}

    As evidenced in \cref{fig:bias_spatial_distribution}, an important component of the overall accuracy of ABC is the reduction of the systematic bias introduced by  dynamical model deficiencies.
    \cref{fig:bias_spatial_distribution} presents the spatial distribution of this bias by plotting the average difference between forecasts and observations over all forecast dates. 
    The precipitation maps reveal 
    a wet bias over the northern half of the \US for \cfs (average bias: 8.32 mm)
    and a dry bias over the south-east part of the \US for \ecmwf (average bias: $-$8.12 mm).
    In this case, ABC eliminates the \cfs wet bias (average bias: $-$0.46 mm) 
    and slightly alleviates the \ecmwf dry bias (average bias: $-$6.24 mm). 
    For temperature, 
    we observe a cold bias over the eastern half of the \US for \cfs (average bias: $-$1.2\degree C) and notice a mixed pattern of cold and warm biases over the western half of the U.S for \ecmwf (average bias: $-$0.30\degree C). In this case, although ABC does not eliminate these biases entirely, it reduces the magnitude of the cold eastern bias by bringing \cfs forecasts closer to observations (average bias: $-0.18\degree\textup{C}$) and reduces the mixed \ecmwf bias (average bias: $-0.04\degree\textup{C}$).
    
    We observe comparable bias reductions when ABC is applied to the SubX multimodel mean in \opt{submit}{Figure~S11}\opt{arxiv}{\cref{fig:bias_spatial_distribution_subx_mean}},
    improved dampening of bias relative to quantile mapping, LOESS, and neural network baselines in \opt{submit}{Figure~S12}\opt{arxiv}{\cref{fig:bias_spatial_distribution_cfsv2}},
    and improved dampening of bias relative to operationally debiased temperature and CFSv2  precipitation in \opt{submit}{Figure~S13}\opt{arxiv}{\cref{fig:bias_spatial_distribution_deb_abc}}.
    Since each bias correction is based on historical data and weather is non-stationary, the  remaining residual bias patterns may be indicative of recent regional shifts in average temperature or precipitation, e.g., decreased average precipitation in the Southeastern U.S.\ or increased average temperature on the country's coasts.

     \newcommand{\cscaption}[4]{
        (\textbf{a}) To summarize the impact of #1 on \aecmwf skill improvement for precipitation weeks 3-4, 
        we divide our forecasts into 10 bins, determined by the deciles of #1, 
        and display above each bin map the probability of positive impact in each bin along with a 95\% bootstrap confidence interval. The highest probability of positive impact is shown in blue, and the lowest probability of positive impact is shown in red. We find that #1 is most likely to have a positive impact on skill improvement in #2 and least likely in #3. (\textbf{b}) The forecast most impacted by #1 in #4.
        }
        \newcommand{\varname}{\texttt{hgt\_500\_pc1}} 
        \begin{figure}[th!]
        \centering
        \includegraphics[width=\textwidth, valign=t]{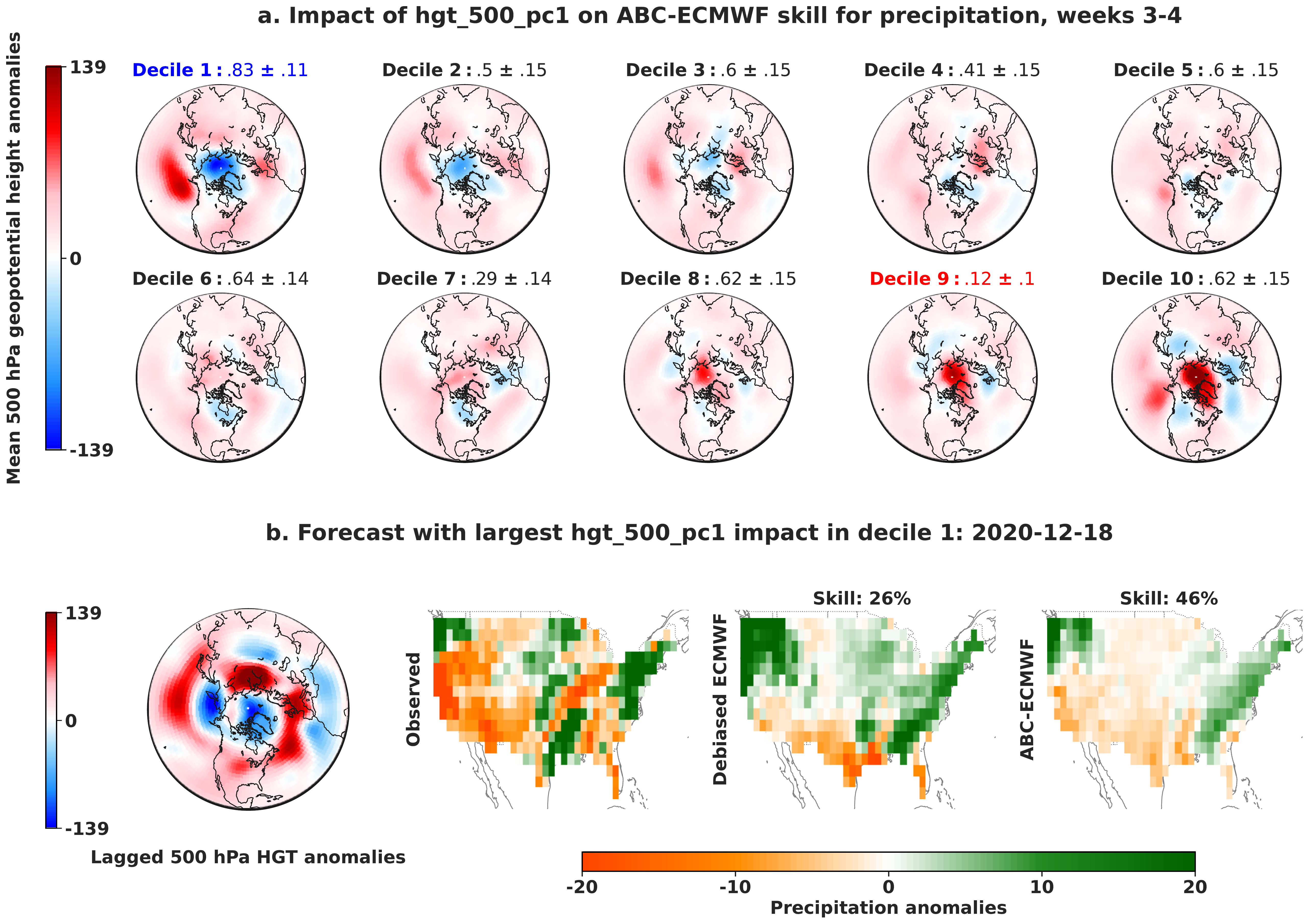}
        \caption{
        \textbf{Impact of the first 500 hPa geopotential heights principal component (\varname) on adaptive bias correction (ABC) %
        skill improvement.}
        \cscaption{\varname}{decile 1, which features a positive Arctic Oscillation (AO) pattern,}{decile 9, which features AO in the opposite phase}{decile 1 is also preceded by a positive AO pattern and replaces the wet debiased \ecmwf forecast with a more skillful dry pattern in the west}}
        \label{fig:shapley_effects_hgt_500_eof1_vs}
        \end{figure}

        \begin{figure}[th!]
        \centering
        \includegraphics[width=\textwidth, %
                         height = 0.7\textheight, 
                         keepaspectratio]{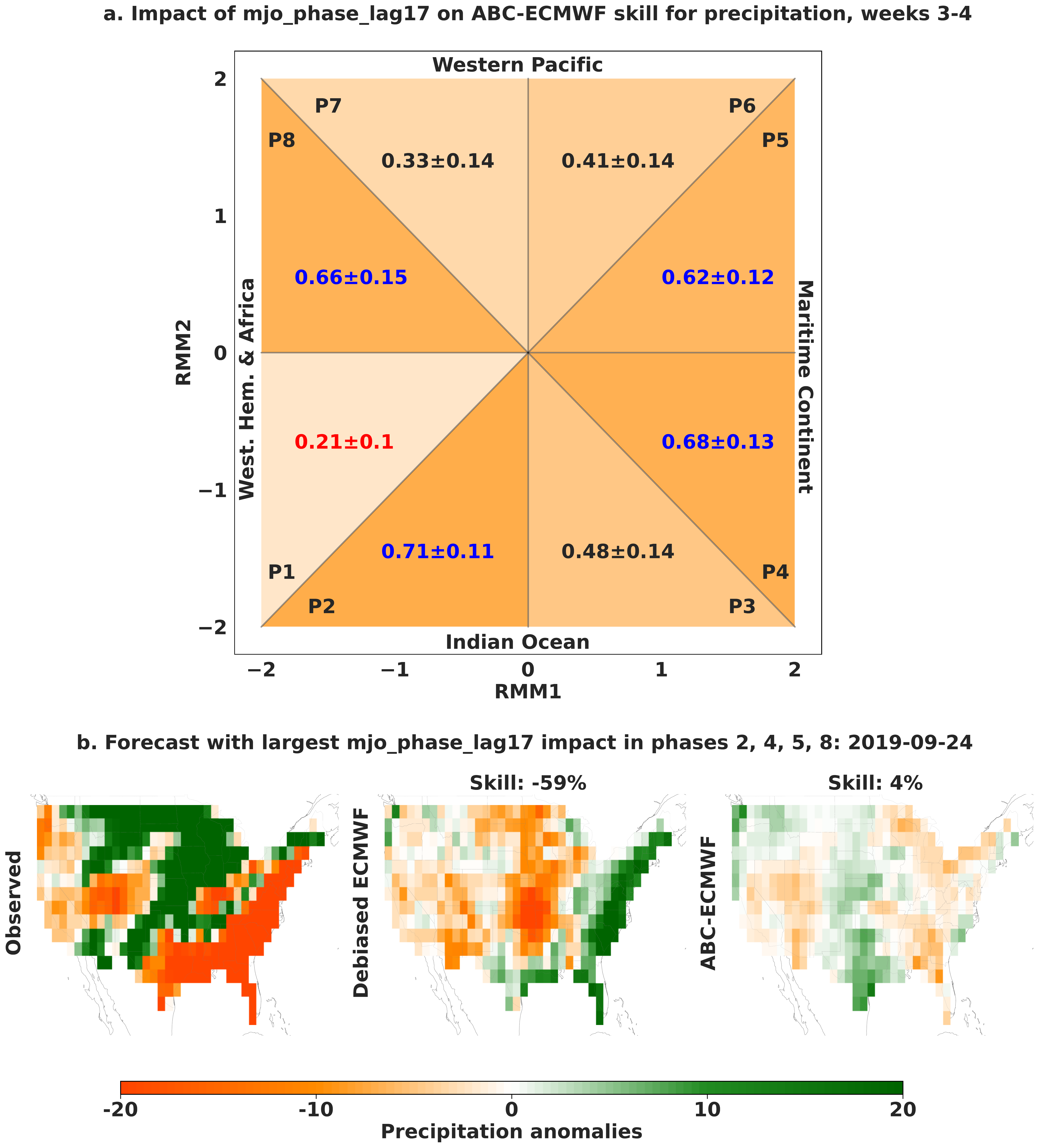}
        \caption{\textbf{Impact of the Madden-Julian Oscillation phase (\texttt{mjo\_phase}) on adaptive bias correction (ABC) %
        skill improvement.} (\textbf{a}) To summarize the impact of \texttt{mjo\_phase} on \aecmwf skill improvement for precipitation weeks 3-4, we compute the probability of positive impact and an associated 95\% bootstrap confidence interval in each lagged MJO phase bin and adopt the methodology of~\cite{wheeler2004all} to create an MJO phase space diagram. 
        The highest probabilities of positive impact (those falling within the confidence interval of the highest probability overall) are shown in blue and the lowest probability of positive impact is shown in red.
        We find that positive impact on skill improvement is most common in phases 2, 4, 5, and 8 and least common in phase 1. (\textbf{b}) The forecast most impacted by \texttt{mjo\_phase} in phases 2, 4, 5, and 8 avoids the strongly negative skill of the debiased \ecmwf baseline.}
        \label{fig:shapley_effects_mjo2_vs}
        \end{figure}

    \subsection*{Identifying statistical forecasts of opportunity}
    
    The results presented so far evaluate overall model skill, averaged across all forecast dates. 
    However, there is a growing appreciation that subseasonal forecasts can benefit from selective deployment during ``windows of opportunity,'' periods defined by observable climate conditions in which specific forecasters are likely to have higher skill~\citep{mariotti2020windows}.
    In this section, we propose a practical 
    \emph{opportunistic ABC workflow}
    that uses a candidate set of explanatory variables to identify windows 
    in which ABC is especially likely to improve upon a baseline model.
    The same workflow can
    be used to explain the skill improvements achieved by ABC in terms of the explanatory meteorological variables.
    
    The opportunistic ABC workflow 
    is based on the equitable credit assignment principle of Shapley~\cite{shapley1953value} and measures the impact of explanatory variables on individual forecasts using Cohort Shapley~\citep{mase2019explaining} and overall variable importance using Shapley effects~\citep{song2016shapley} (see \cref{sec:methods} for more details). 
    We use these Shapley measures to determine the contexts in which ABC offers improvements, in terms of climate variables with known relevance for subseasonal forecasting accuracy. 
        As a running example, we use 
        our workflow
        to explain the skill differences between \aecmwf and debiased \ecmwf when predicting precipitation in weeks 3-4. As our candidate explanatory variables we use 
        Northern Hemisphere geopotential heights (HGT) at 500 and 10 hPa, the phase of the Madden-Julian Oscillation (MJO), Northern Hemisphere sea ice concentration (ICEC), global sea surface temperatures (SST), the multivariate El \Nino-Southern Oscillation index~(MEI.v2) \citep{wolter1993monitoring}, and the target month. 
        All variables are lagged as described in \cref{sec:methods} to ensure that they are observable on the forecast issuance date.
        
        We first use Shapley effects to determine the overall importance of each variable in explaining the precipitation skill improvements of \aecmwf. 
        As shown in \opt{submit}{Figure~S14}\opt{arxiv}{\cref{fig:shapley_effects}}, the most important explanatory variables are
        the first two principal components (PCs) of 500 hPa geopotential height, 
        the MJO phase,  
        the second PC of 10 hPa geopotential height, and
        the first PC of sea ice concentration.
        These variables are consistent with the literature exploring the dominant contributions to subseasonal precipitation. 
        The 500 hPa geopotential height plays a crucial role in conveying information about the thermal structure of the atmosphere and indicates synoptic circulation changes~\citep{christidis2015changes}. 
        The MJO phase influences weather and climate phenomena within both the tropics and extratropics, resulting in a global influence of MJO in modulating temperature and precipitation~\citep{woolnough2019madden}. 
        The 10 hPa geopotential height is a known indicator of polar vortex variability leading to lagged impacts on sea level pressure, surface temperature, and precipitation~\citep{merryfield2020current}. 
        Finally, sea ice concentration has a strong impact on surface turbulent heat fluxes and therefore near-surface temperatures~\citep{chevallier2019role}. 
        
        We next use Cohort Shapley to identify the contexts in which each variable has the greatest impact on skill. 
        For example, \cref{fig:shapley_effects_hgt_500_eof1_vs} summarizes the impact of the first 500 hPa geopotential heights PC 
        (\texttt{hgt\_500\_pc1}) 
        on \aecmwf skill improvement. This display divides our forecasts into 10 bins, determined by the deciles of \texttt{hgt\_500\_pc1}, and computes the probability of positive impact in each bin. We find that \texttt{hgt\_500\_pc1} 
        is most likely to have a positive impact impact on skill improvement in decile 1, which features a positive Arctic Oscillation (AO) pattern, and least likely in decile 9, which features AO in the opposite phase. The \aecmwf forecast most impacted by \texttt{hgt\_500\_pc1}  
        in decile 1 is also preceded by a positive AO pattern and replaces the wet debiased \ecmwf forecast with a more skillful dry pattern in the west. 
        Similarly, \cref{fig:shapley_effects_mjo2_vs} summarizes the impact of the  MJO phase 
        (\texttt{mjo\_phase}) on \aecmwf skill improvement. 
        Importantly, while skill improvement is sometimes achieved with an especially skillful ABC forecast (as in \cref{fig:shapley_effects_hgt_500_eof1_vs}) it can also be achieved by recovering from an especially poor baseline forecast.  The latter is what we see at the bottom of \cref{fig:shapley_effects_mjo2_vs}, where the highest impact ABC forecast avoids the strongly negative skill of the baseline debiased \ecmwf forecast. 
        
\begin{figure}[t!]
        \resizebox{\textwidth}{!}{
        \includegraphics{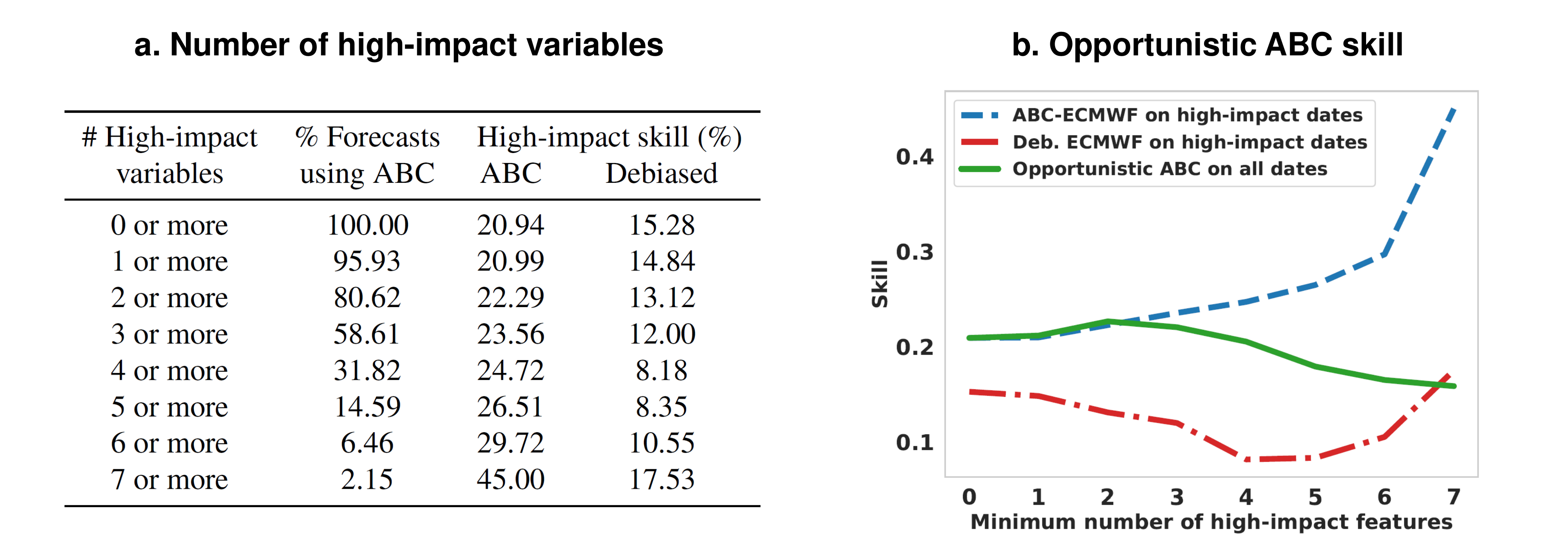}
        }
        \caption{\textbf{Defining windows of opportunity for opportunistic adaptive bias correction (ABC) forecasting.}
        Here we focus on forecasting precipitation in weeks 3-4. 
        (\textbf{a}) When more explanatory variables fall into high-impact deciles or bins (e.g., the blue bins of  \cref{fig:shapley_effects_hgt_500_eof1_vs,fig:shapley_effects_mjo2_vs}), the mean skill of \aecmwf improves, but the percentage of forecasts using ABC declines.
        (\textbf{b}) The overall skill of opportunistic ABC is maximized when \aecmwf is deployed for target dates with two or more high-impact variables and standard debiased \ecmwf is deployed otherwise.}
        \label{fig:opportunity}
        \end{figure}

\begin{figure}[t!]
        \centering
        \includegraphics[width=\textwidth,height = 0.89\textheight, keepaspectratio]{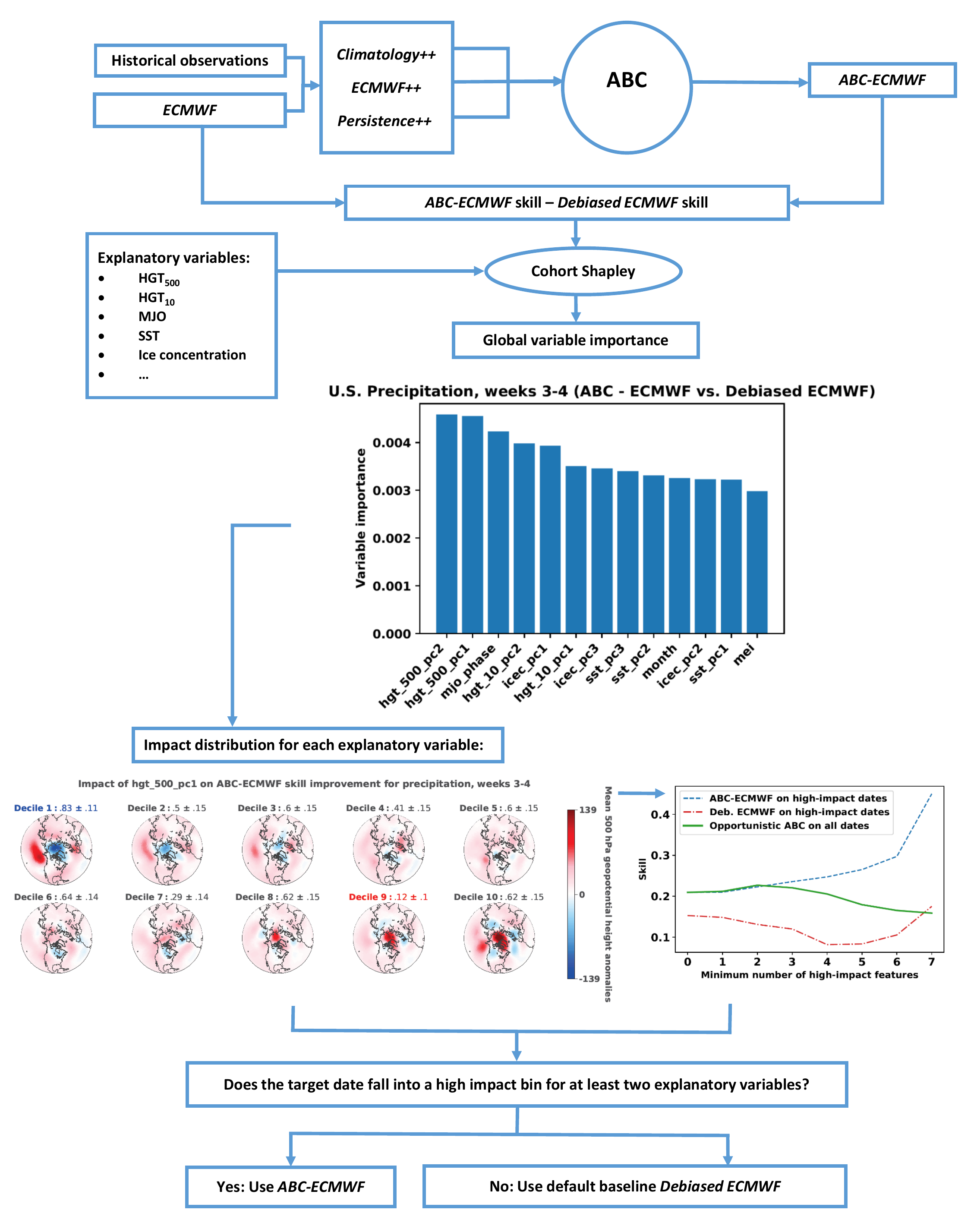}
        \caption{\textbf{Schematic of the opportunistic adaptive bias correction (ABC) workflow.} Opportunistic ABC uses historical ABC and baseline forecasts and a candidate set of explanatory variables to identify windows of opportunity for selective deployment of ABC in an operational setting.}
        \label{fig:abc_flowchart}
        \end{figure}

        Finally, we use the identified contexts to define windows of opportunity for operational deployment of ABC.
        Indeed, since all explanatory variables are observable on the forecast issuance date, one can 
        selectively apply ABC  
        when multiple variables are likely to have a positive impact on skill
        and otherwise issue a default, standard forecast (e.g., debiased \ecmwf). 
        We call this selective forecasting model \emph{opportunistic ABC}.
        How many high-impact variables should we require when defining these windows of opportunity? 
        We say a variable is ``high-impact'' if the positive impact probability for its decile or bin is within the confidence interval of the highest probability overall.
        Requiring a larger number of high-impact variables will tend to increase the skill gains of ABC but simultaneously reduce the number of dates on which ABC is deployed.
        \cref{fig:opportunity} illustrates this trade-off for \aecmwf and shows that opportunistic ABC skill is maximized when two or more high-impact variables are required.
        With this choice, ABC is used for approximately $81\%$ of forecasts and debiased \ecmwf is used for the remainder. 
        \cref{fig:abc_flowchart} 
        summarizes the complete opportunistic ABC workflow, from the identification of windows of opportunity through the selective deployment of either ABC or a default baseline forecast for a given target date. 
        
\section{Discussion} \label{sec:discussion}
    Dynamical models have shown increasing skill in accurately forecasting the weather~\citep{bauer2015quiet}, but they still contain systematic biases that compound on subseasonal time scales and suppress forecast skill~\citep{flato2014evaluation, zadra2018systematic, zhang2021evaluation,dutra2021late}.
    ABC learns to correct these biases by adaptively integrating dynamical forecasts, historical observations, and recent weather trends.
    When applied to the leading subseasonal model from \ecmwf, ABC improves forecast skill by 60-90\% (over baseline skills of 0.18-0.25) for precipitation and 40-69\% (over baseline skills of 0.11-0.15) for temperature.
    The same approach 
    substantially reduces the 
    forecasting errors of 
    seven additional operational subseasonal forecasting models as well as their multimodel mean,
    with less skillful input models performing nearly as well as the ECMWF model after applying the ABC correction.
    This finding suggests that systematic errors in dynamical models are a primary contributor to observed skill differences and that ABC provides an effective mechanism for reducing these heterogeneous errors.   
    Because ABC is also simple to implement and deploy in real-time operational settings, adaptive bias correction represents a computationally inexpensive strategy for upgrading operational models, while conserving valuable human resources. 

    While the learned correction of systematic errors can play an important role in skill improvement, it is no substitute for scientific improvements in our understanding and representation of the processes underlying subseasonal predictability.
    As such, we view ABC as a complement for improved dynamical model development.
    Fortunately, ABC is designed to be adaptive to model changes. 
    As operational models are upgraded, process models improve, and systematic biases evolve, our ABC training protocol is designed to ingest the upgraded model forecasts and hindcasts reflecting those changes. 

    To capitalize on higher-skill forecasts of opportunity, 
    we have also introduced an opportunistic ABC workflow that explains the skill improvements of ABC in terms of a candidate set of environmental variables, 
    identifies high-probability windows of opportunity based on those variables, and selectively deploys either ABC or a baseline forecast to maximize expected skill.
    The same workflow can be applied to explain the skill improvements of any forecasting model and, unlike other popular explanation tools \citep[e.g.,][]{ribeiro2016should,lundberg2017unified}, avoids expensive model retraining, requires no generation of additional forecasts beyond those routinely generated for operational or hindcast use, and allows for explanations in terms of variables that were not explicitly used in training the model.

    Overall, we find that correcting dynamical forecasts using ABC yields an effective and scalable strategy to optimize the skill of the next generation of subseasonal forecasting models. 
    We anticipate 
    that our hybrid dynamical-learning framework 
    will benefit both research and 
    operations, and we release our open-source code 
    to facilitate future adoption and development.

\section{Methods} \label{sec:methods}
\subsection*{Dataset}
All data used in this work was obtained from the \dataset dataset \citep{subseasonalusa2021}.
The spatial variables were interpolated onto a $1.5^\circ\times 1.5^\circ$ latitude-longitude grid, and all daily observations (with two exceptions noted below) were aggregated into two-week moving averages. 
As ground-truth measurements, we extracted daily gridded observations of average $2$-meter temperature in $^\circ$C~\citep{fan2008global} and precipitation in mm~\citep{xie2007gauge,chen2008assessing,xie2010cpc}.
For our explanatory variables, we obtained the daily PCs of 10 hPa and 500 hPa stratospheric geopotential height~\citep{kalnay1996ncep} extracted from global 1948-2010 loadings, the daily PCs of sea surface temperature and sea ice concentration~\citep{reynolds2007daily} using global 1981-2010 loadings, the daily MJO phase~\citep{wheeler2004all}, and the bimonthly MEI.v2~\citep{wolter1993monitoring,wolter1998measuring,wolter2011nino}.
Precipitation was summed over two-week periods, and the MJO phase was not aggregated.
Finally, we extracted twice-weekly ensemble mean forecasts of temperature and precipitation from the ECMWF S2S dynamical model~\citep{vitart2012subseasonal} and ensemble mean forecasts of temperature and precipitation from seven models participating in the SubX project, including five coupled atmosphere-ocean-land dynamical models (NCEP-CFSv2, GMAO-GEOS, NRL-NESM, RSMAS-CCSM4, ESRL-FI) and two models with atmosphere and land components forced with prescribed sea surface temperatures (EMC-GEFS, ECCC-GEM)~\citep{pegion2019subseasonal}. 
The SubX multimodel mean forecast was obtained by calculating, for each target date, the mean prediction over all available SubX models using the most recent forecast available from each model within a lookback window of size equal to six days. Each candidate SubX model is represented by its ensemble mean forecast. Two sets of candidate models are considered. When calculating SubX ensemble mean for weeks 1-2 and weeks 3-4, we consider NCEP-CFSv2, GMAO-GEOS, NRL-NESM, RSMAS-CCSM4, ESRL-FI, EMC-GEFS, and ECCC-GEM. When generating mean forecast for weeks 5-6, we consider NCEP-CFSv2, GMAO-GEOS, NRL-NESM, and RSMAS-CCSM4 only, as the remaining models do not produce forecast data for weeks 5-6.

\subsection*{Forecasting tasks and skill assessment} 
    We consider two prediction targets: 
    average temperature ($^\circ$C) and accumulated precipitation (mm) over a two-week period. These variables are forecasted at two time horizons: 15-28 days ahead (weeks 3-4) and 29-42 days ahead (weeks 5-6). 
    We forecast each variable at $G=376$ 
    grid points on a $1.5^\circ \times 1.5^\circ$ grid across 
    the contiguous U.S., bounded by latitudes 25N to 50N and longitudes 125W to 67W.
    To provide the most realistic assessment of forecasting skill  \citep{risbey2021standard}, all predictions in this study are formed in a real forecast manner that mimics operational use. In particular, to produce a forecast for a given target date, all learning-based models are trained and tuned only on data observable on the corresponding forecast issuance date.

    For evaluation, we adopt the exact protocol of the recent Subseasonal Climate Forecast Rodeo competition, run by the U.S.\ Bureau of Reclamation in partnership with the National Oceanic and Atmospheric Administration, U.S.\ Geological Survey, U.S.\ Army Corps of Engineers, and California Department of Water Resources \citep{nowak2017sub}.
    In particular, for a two-week period starting on date $t$, let $\gt_t \in \R^G$ denote the vector of ground-truth measurements $y_{t,g}$ for each grid point $g$ and $\predgt_t \in \R^G$ denote a corresponding vector of forecasts. 
    In addition, define climatology $\mathbf{c}_{t}$ 
    as the average ground-truth values for a given month and day over the years 1981-2010.
    We evaluate each forecast using uncentered anomaly correlation skill  \citep{wilks2011statistical,nowak2017sub}, 
    \begin{align}\textstyle
        \skill(\predgt_t, \gt_t) 
        = \frac{\langle \predgt_t - \mbf{c}_t, \gt_t - \mbf{c}_t\rangle}{\|\predgt_t - \mbf{c}_t\|_2 \cdot \|\gt_t - \mbf{c}_t\|_2} 
        \in [-1,1], 
    \end{align}
    with a larger value indicating higher quality.
    For a collection of target dates, we report average skill using progressive validation~\citep{blum1999beating} to mimic operational use. 
    For a fixed grid point, we define spatial skill analogously as the uncentered anomaly correlation across all target dates in the evaluation period.
    
\subsection*{Operational \ecmwf, \cfs, and SubX debiasing}
We bias correct a uniformly-weighted ensemble of the ECMWF control forecast and its 50 ensemble forecasts following the ECMWF operational protocol~\citep{ecmwf2022reforecast}: 
for each target forecast date and grid point, we bias correct the 51-member ensemble forecast by subtracting the equal-weighted 11-member ECMWF ensemble reforecast averaged over all dates from the last $20$ years within $\pm6$ days from the target month-day combination
and then adding the average ground-truth measurement over the same dates.

Following \cite{nowak2017sub}, we bias correct a uniformly-weighted 32-member \cfs ensemble forecast, formed from four model initializations averaged over the eight most recent $6$-hourly issuances, in the following way: 
for each target forecast date and grid point, we bias correct the 32-member ensemble forecast by subtracting 
the equal-weighted 8-member \cfs ensemble hindcast averaged over all dates from 1999 to 2010 inclusive matching the target day and month 
and then adding the average ground-truth measurement over the same dates.
We bias correct the SubX multimodel mean forecast in an identical manner, using the SubX multimodel mean reforecasts from 1999 to 2010 inclusive. 
\subsection*{Adaptive bias correction} \label{sec:models_details}
    ABC is a uniformly-weighted ensemble of three machine learning models, \nwppp, \climpp, and \perpp, detailed below.  
    \opt{submit}{A schematic of ABC model input and output data can be found in Figure~S15, and supplementary algorithm details can be found in Supplementary Methods.}\opt{arxiv}{A schematic of ABC model input and output data can be found in \cref{fig:abc_schematic}, and supplementary algorithm details can be found in \nameref{sec:supp_methods}.}

    \newcommand{\nwpppalg}{\opt{submit}{Algorithm~S1}\opt{arxiv}{\cref{alg:nwppp}}}
    \tbf{\nwppp} (\nwpppalg) is a three-step approach to \nwp model correction: 
    (i) adaptively select a window of observations around the target day of year and a range of issuance dates and lead times for ensembling based on recent historical performance,
    (ii) form an ensemble mean forecast by averaging over the selected range of issuance dates and lead times, and
    (iii) bias-correct the ensemble forecast for each site by adding the mean value of the target variable and subtracting the mean forecast over the selected window of observations.
    Unlike standard debiasing strategies, which employ static ensembling and bias correction, \nwppp adapts to heterogeneity in forecasting error by learning to vary the amount of ensembling and the size of the observation window over time.

    For a given target date $\tstar$ and lead time $\lstar$, the \nwppp training set $\trainset$ is restricted to data fully observable one day prior to the issuance date, that is, to dates $t \leq \tstar - \lstar - L - 1$ where $L = 14$ represents the forecast period length.
    For each target date, \nwppp is run with the hyperparameter configuration that achieved the smallest mean progressive geographic root mean squared error (RMSE) over the preceding $3$ years. 
    Here, \emph{progressive} indicates that each candidate model forecast is generated using all training data observable prior to the associated forecast issuance date. 
    Every configuration with \emph{span} $s \in \{ 0, 14, 28, 35 \}$ (the span is the number of days included on each side of the target day of year), number of averaged issuance dates $\dstar \in \{1, 7, 14, 28, 42\}$, and leads $\leads = \{29\}$ for the weeks 5-6 lead time and $\leads \in \{ \{15\}, [15,22], [0,29], \{29\} \}$ the weeks 3-4 lead time was considered.

    \newcommand{\climppalg}{\opt{submit}{Algorithm~S2}\opt{arxiv}{\cref{alg:climpp}}}
    Inspired by climatology, \textbf{\climpp} (\climppalg) makes no use of the dynamical forecast and rather outputs the historical geographic median (if the user-supplied loss function is RMSE) or mean (if loss $=$ MSE) of its target variables over all days in a window around the target day of year. Unlike a static climatology, \climpp adapts to target variable heterogeneity by learning to vary the size of the observation window and the number of training years over time.

    For a given target date $\tstar$ and lead time $\lstar$, the \climpp training set $\trainset$ is restricted to data fully observable one day prior to the issuance date, that is, to dates $t \leq \tstar - \lstar - L - 1$ where $L = 14$ represents the forecast period length.
    For each target date, \climpp is run with the hyperparameter configuration 
    that achieved the  smallest mean progressive geographic RMSE over the preceding $3$ years.
    All spans $s \in \{0, 1, 7, 10\}$ were considered.
    All precipitation configurations used the geographic MSE loss and all available training years.
    All temperature configurations used the geographic RMSE loss and either all available training years or $Y=29$.
    For shorter than subseasonal lead times (e.g., weeks 1-2), \climpp is excluded from the ABC forecast and only \nwppp and \perpp are averaged.

    \newcommand{\perppalg}{\opt{submit}{Algorithm~S3}\opt{arxiv}{\cref{alg:perpp}}}
    \textbf{\perpp} (\perppalg) accounts for recent weather trends by fitting an ordinary least-squares regression per grid point to optimally 
    combine lagged temperature or precipitation measurements, climatology, and a \nwp ensemble forecast.
    For a given target date $\tstar$ and lead time $\lstar$, the \perpp training set $\trainset$ is restricted to data fully observable one day prior to the issuance date, that is, to dates $t \leq \tstar - \lstar - L - 1$ where $L = 14$ represents the forecast period length.
    In \perppalg, the set $\leads$ represents the full set of subseasonal lead times available in the dataset, i.e., $\leads = [0,29]$.

\subsection*{Debiasing baselines} 
\textbf{NN-A}~\citep{fan2021using} learns a non-linear mapping between daily corrected \cfs precipitation and temperature and observed precipitation and temperature for the contiguous U.S. In particular, the model's inputs (predictors) consist of \cfs bias-corrected ensemble mean for total precipitation and temperature anomalies as well as the observed climatologies for precipitation and temperature. The model's target variables (predictands) are observed temperature anomalies and total precipitation. Both daily target variables are converted to two-weekly mean and two-weekly total and are predicted simultaneously for the entire forecast domain.
NN-A is a neural network with a single hidden layer consisting of $K=200$ hidden neurons.
This architecture enables the model to account for both non-linear relationships among input and target variables as well as their spatial dependency and the co-variability that characterize these variables. 
For a given lead time, the NN-A model was trained on all available data from January 2000 to December 2017 (inclusive) save for those dates that were unobservable on the issuance date associated with a January 1, 2018 target date.
Each NN-A model was trained using the Adam algorithm~\citep{kingma2014adam} for 10001 epochs without dropout as in~\citep{fan2021using} and used relu activations and the default batch size (32) and learning rate (0.001) from Tensorflow~\citep{tensorflow2015-whitepaper}.

\textbf{LOESS debiasing} \citep{cleveland1988locally} adds a correction to a dynamical model forecast using locally estimated scatterplot smoothing. Using all dates prior to 2018 with available ground-truth measurements and (re)forecast data, the measurements for each month-day combination (save February 29) are averaged, resulting in a sequence of $365$ values. The same is done for the forecasts. A local linear regression is run on each of these sequences using a fraction of $0.1$ of the points to fit each value. The end result is two smoothed sequences of $365$ values, one with the measurement data and the other one with forecast data. The entrywise difference (in the case of temperature) or ratio (in the case of precipitation) between these sequences is used as a correction to be added (in the case of temperature) or multiplied (in the case of precipitation) to the forecasts made in 2018 and beyond, based on the target forecast day and month. Note that the locality of the smoothed corrections, which only use consecutive days in the calendar year and do not wrap around from December to January, ensures that every forecast is made using only training data observable on the forecast issuance date.

\textbf{Quantile mapping} \citep{panofsky1968some} corrects a base dynamical forecast by aligning the quantiles of forecast and measurement data.  For our training set, we use all dates prior to 2018 with available ground-truth measurements and (re)forecast data. 
For a given grid point, target date, and dynamical model forecast, we first identify the quantile rank of the forecasted value amongst all training set forecasts issued for the same month-day combination.  
If the quantile rank exceeds $90\%$, we replace its value with $90\%$; if the quantile rank falls below $10\%$, we replace its value with $10\%$. 
We then add to the forecast the corresponding quantile of the training set measurements for the target month and day and subtract the corresponding quantile of the training set forecasts for the target month and day. In the case of precipitation, if the resulting value is negative, we set the forecast to zero. 
\subsection*{Cohort Shapley and Shapley effects}

Cohort Shapley and Shapley effects use \emph{Shapley values} to quantify the impact of variables on outcomes.  Shapley values are based on work in game theory~\citep{shapley1953value} exploring how to assign appropriate rewards to individuals who contribute to an outcome as part of a team.  When applied to explanatory variables, Shapley values can be thought of as roughly analogous to the coefficients in a linear regression.  Importantly, unlike linear regression coefficients, Shapley values are applicable in settings where the interaction among variables is highly non-linear.  The procedure for computing Shapley values involves testing how much a change to one explanatory variable influences a target outcome.  These tests are carried out by measuring how the target outcome varies when a given explanatory variable changes in the context of subsets of other explanatory variables.  

Shapley effects~\citep{song2016shapley} 
are a specific instantiation of the general Shapley value principle, designed 
for measuring variable importance.
For a given outcome variable to be explained (for example, the skill difference between ABC-ECMWF and operationally-debiased ECMWF measured on each forecast date) and a collection of candidate explanatory variables (for example, relevant meteorological variables observed at the time of each forecast's issuance), the Shapley effects are overall measures of variable importance that quantify how much of the outcome variable's variance is explained by each candidate explanatory variable.  
Cohort Shapley values~\citep{mase2019explaining} provide a more granular application of Shapley values by quantifying the impact of each explanatory variable on the measured outcome of each individual forecast.  

\subsection*{Opportunistic ABC workflow} \label{sec:interpretability}

Here we detail the steps of the opportunistic ABC workflow illustrated in \cref{fig:abc_flowchart} using \ecmwf as an example dynamical input.  The same workflow applies to any other \nwp input.

\begin{enumerate}[leftmargin=0.5cm]
    \item Identify a set of $V$ candidate explanatory variables. Here we use the 
    temporal variables 
    enumerated in \cite[Fig.~2]{hwang2019improving} augmented with the first two PCs of 500 hPa geopotential heights and the target month.  To ensure that the workflow can be deployed operationally, we use lagged observations with lags chosen so that each variable is observable on the forecast issuance date. MEI.v2 is lagged by $45$ days when forecasting weeks 3-4 and by $59$ days for weeks 5-6. The other variables are lagged by $30$ days for weeks 3-4 and by $44$ days for weeks 5-6.
    \item Compute the temporal skill difference between \aecmwf and debiased \ecmwf for each target date in the evaluation period.
    \item For each continuous explanatory variable (e.g., hgt\_500\_pc2), divide the evaluation period forecasts into 10 bins, determined by the deciles of the explanatory variable.  For each categorical variable (e.g., mjo\_phase), divide the forecasts into bins determined by the categories (e.g., MJO phases).
    \item Use the \texttt{cohortshapley} Python package to compute overall variable importance (measured by Shapley effects) and forecast-specific variable impact values explaining the skill differences.
    \item Within each variable bin, compute the fraction of forecasts with positive Cohort Shapley impact values. Report that fraction as an estimate of the probability of  positive variable impact, and compute a 95\% bootstrap confidence interval. 
    Flag all bin probabilities within the confidence interval of the highest probability bin as high impact; similarly, flag all bin probabilities within the confidence interval of the lowest probability bin as low impact. 
    The remaining bins -- those that fall outside of both confidence intervals -- have intermediate impact and are not flagged as either low or high impact.
    Visualize and interpret the highest and lowest impact bins.
    \item Identify the forecast most impacted by the explanatory variable in the high impact bins. Visualize the  \aecmwf and debiased \ecmwf forecasts and the associated explanatory variable for that target date.
    \item For each $k \in \{0,\dots,V\}$, compute opportunistic ABC skill when $k$ or more explanatory variables fall into high impact bins.  Let $k^\star$ represent the integer at which opportunistic ABC skill is maximized.
    \item At each future forecast issuance date, deploy \aecmwf if $k^\star$ or more explanatory variables fall into high impact bins and deploy debiased \ecmwf otherwise.
\end{enumerate}
\section*{Data availability}

The \dataset dataset used in this study has been deposited on Microsoft Azure and is available for download via the \datapackage Python package: \url{https://github.com/microsoft/subseasonal_data}. Source data
are provided with this paper. This work is based on S2S data. S2S is a joint initiative of the World Weather Research Programme (WWRP) and the World Climate Research Programme (WCRP). The original S2S database is hosted at ECMWF as an extension of the TIGGE database. We acknowledge the agencies that support the SubX system, and we thank the climate modeling groups (Environment Canada, NASA, NOAA/NCEP, NRL and University of Miami) for producing and making available their model output. NOAA/MAPP, ONR, NASA, NOAA/NWS jointly provided coordinating support and led development of the SubX system.

\section*{Code availability}
Python 3 code replicating all experiments and analyses in this work is available at
\begin{center}
    \url{https://github.com/microsoft/subseasonal_toolkit}
\end{center}

\def\bibcommenthead{} 
\bibliographystyle{IEEEtran}
\bibliography{references}

\section*{Acknowledgements}
We thank Jessica Hwang for her suggestion to explore Cohort Shapley.
This work was supported by Microsoft AI for Earth (S.M. and G.F.); the Climate Change AI Innovation Grants program (S.M., P.O., G.F., J.C., E.F., and L.M.), hosted by Climate Change AI with the support of the Quadrature Climate Foundation, Schmidt Futures, and the Canada Hub of Future
Earth; 
FAPERJ (Fundação Carlos Chagas Filho de Amparo à Pesquisa do Estado do Rio de Janeiro) grant SEI-260003/001545/2022 (P.O.); 
NOAA grant OAR-WPO-2021-2006592 (G.F., J.C., and L.M.); 
and the National Science Foundation grant PLR-1901352 (J.C.).
\section*{Author contributions}
S.M., P.O., G.F., J.C., E.F., and L.M. designed the research, discussed the results, and contributed to the writing of the manuscript.
S.M., P.O., G.F., M.O., and L.M. performed model runs. S.M., P.O., and L.M. analyzed model outputs and generated figures.
\section*{Competing interests}
The authors declare no competing interests.

\opt{arxiv}{
\clearpage
\appendix
\setcounter{equation}{0}
\renewcommand{\theequation}{S\arabic{equation}}
\pagenumbering{gobble}
\begin{center}
\Large
    Supplementary Information for \\
    Adaptive Bias Correction for Improved Subseasonal Forecasting
\end{center}

\vspace{.5\baselineskip}
\begin{center}
\textbf{Soukayna Mouatadid$^1$,
Paulo Orenstein$^2$,
Genevieve Flaspohler$^{3,4,5}$,
Judah Cohen$^{6,7}$,
Miruna Oprescu$^8$,
Ernest Fraenkel$^9$,
Lester Mackey$^{10}$}
\end{center}

\begin{center}
\footnotesize
$^1${Department of Computer Science, University of Toronto, Toronto, ON, Canada}\\
$^2${Instituto de Matem\'atica Pura e Aplicada, Rio de Janeiro, Brazil}\\
$^3${\textit{n}Line Inc., Berkeley, CA, USA}\\
$^4${Department of Electrical Engineering and Computer Science, Massachusetts Institute of Technology, Cambridge, MA, USA}\\
$^5${Department of Applied Ocean Science and Engineering, Woods Hole Oceanographic Institution, Falmouth, MA, USA}\\
$^6${Atmospheric and Environmental Research, Lexington, MA, USA}\\
$^7${Department of Civil and Environmental Engineering, Massachusetts Institute of Technology, Cambridge, MA, USA}\\
$^8${Department of Computer Science, Cornell University, Ithaca, NY, USA}\\
$^9${Department of Biological Engineering, Massachusetts Institute of Technology, Cambridge, MA, USA}\\
$^{10}${Microsoft Research New England, Cambridge, MA, USA}
\end{center}

\setcounter{figure}{0}
\renewcommand{\figurename}{Figure}
\renewcommand{\thefigure}{S\arabic{figure}}

\section{Supplementary Figures} \label{sec:results_details}
\vspace{-1.73\baselineskip}  
        \begin{figure}[H]
        \centering
        \includegraphics[width=\textwidth,
                         height = 0.85\textheight, 
                         keepaspectratio]{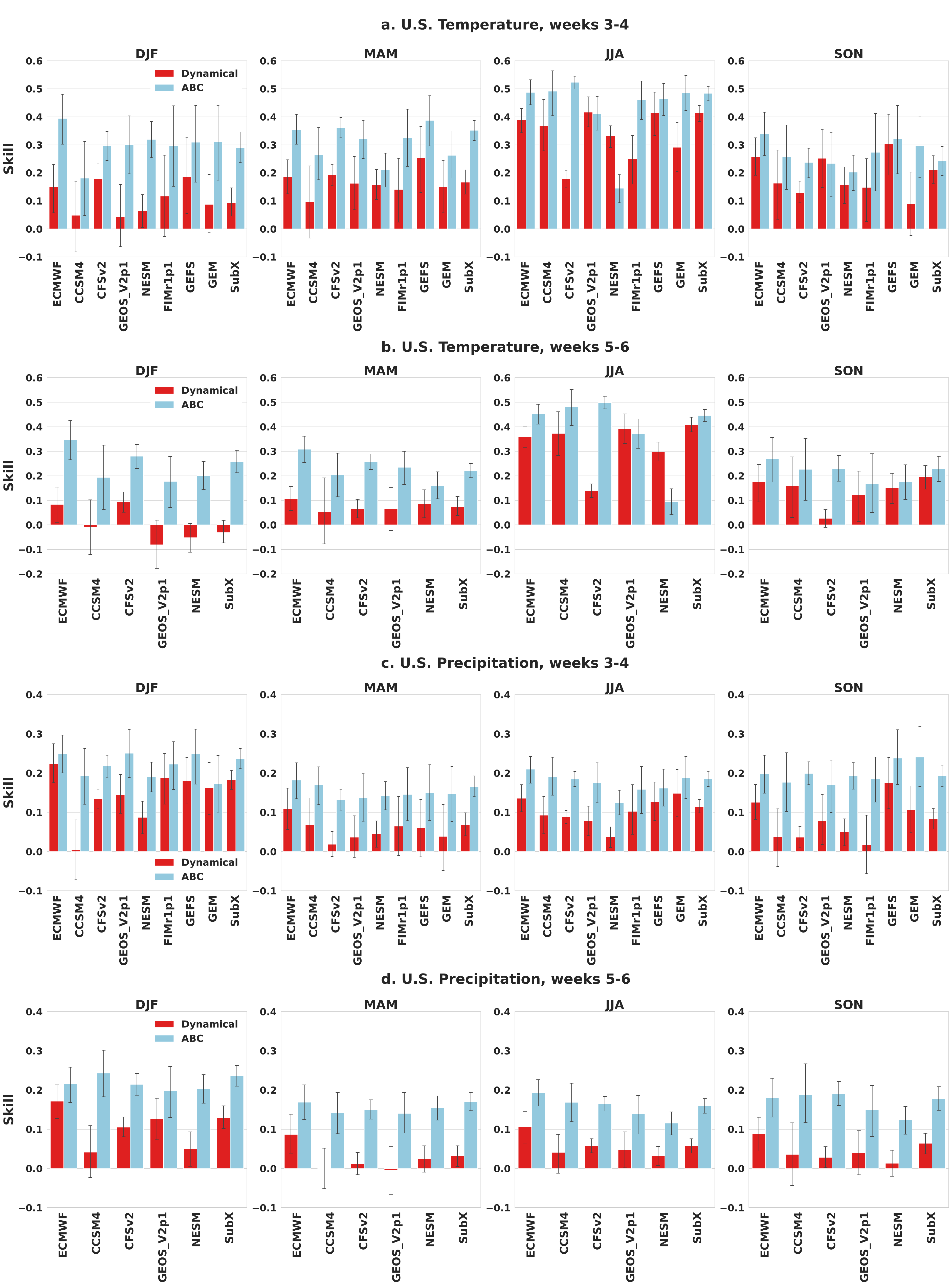}
        \caption{\textbf{Average forecast skill per season for dynamical models (red) and their ABC-corrected counterparts (blue).} 
        For each forecasting task (\textbf{a, b, c, d}), 
        skill is averaged across the contiguous \US and the years 2018--2021 
        with DJF = December, January, February; 
        MAM = March, April, May;
        JJA = June, July, August; and 
        SON = September, October, November.
        The error bars display 95\% bootstrap confidence intervals. Models without forecast data for weeks 5-6 are omitted from the weeks 5-6 panels.}
        \label{fig:barplot_subx_vs_abc_quarterly}
        \end{figure}

        \begin{figure}[h!]
            \centering
            \includegraphics[width=\textwidth]{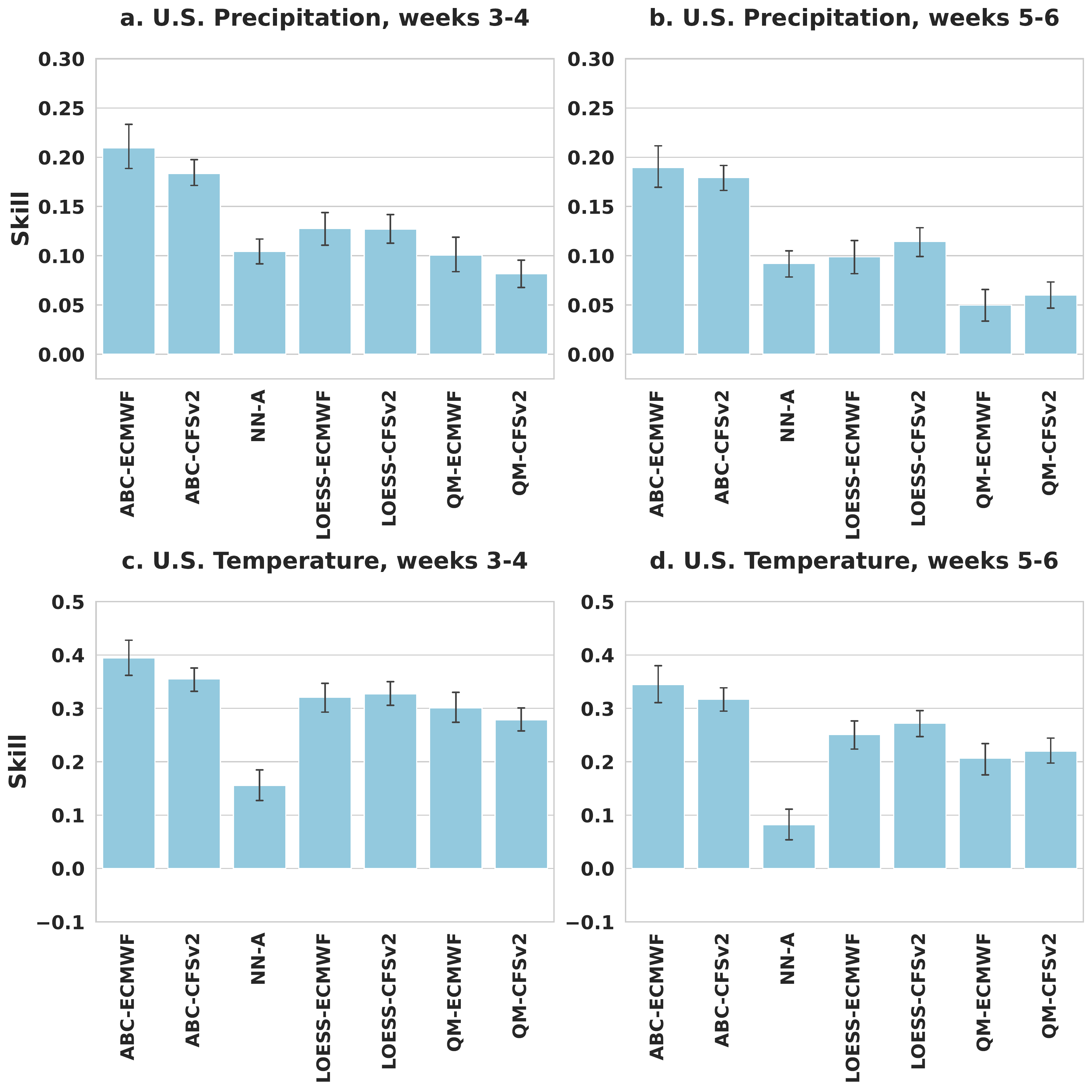}
            \caption{\textbf{Average forecast skill for adaptive bias correction (ABC) and baseline neural network (NN-A),  locally estimated scatterplot smoothing (LOESS), and quantile mapping (QM) corrections of dynamical models.} 
            For each forecasting task (\textbf{a, b, c, d}), 
            skill is averaged across the contiguous \US and the years 2018--2021.
            The error bars display 95\% bootstrap confidence intervals.
            The NN-A correction operates specifically on \cfs model inputs.}
            \label{fig:barplot_nn-a}
        \end{figure} 
        \begin{figure}[h!]
        \centering
        \includegraphics[width=\textwidth]{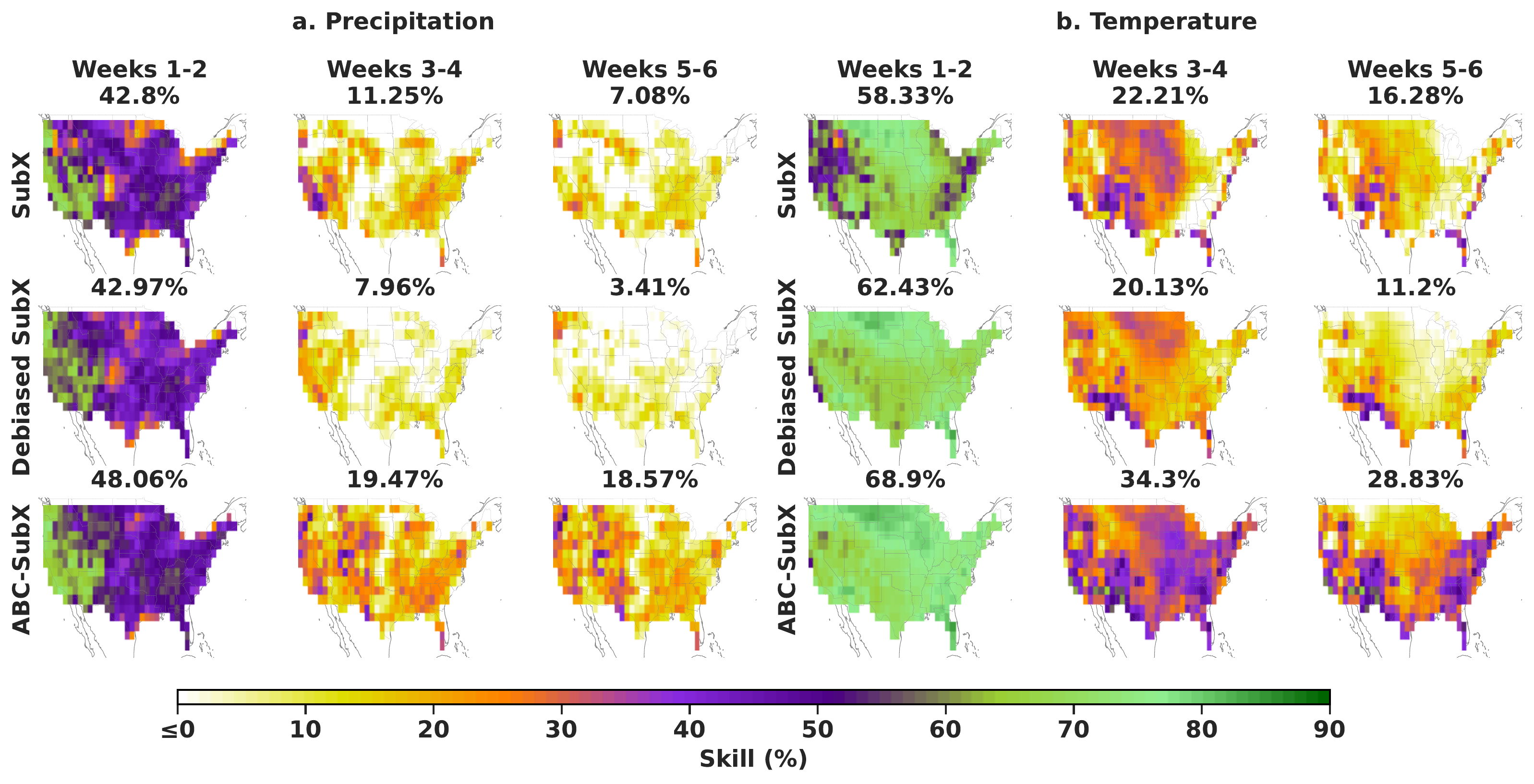}
        \caption{\textbf{Spatial skill distribution of SubX multimodel mean and adaptive bias corrections (ABC).}
        Across the contiguous \US and the years 2018--2021, SubX skill drops precipitously at subseasonal timescales (weeks 3-4 and 5-6), but ABC attenuates the degradation, boosting the skill of the SubX multimodel mean by 109-289\% (over baseline skills of 0.07-0.22).
        Taking the same raw multimodel mean forecasts as input, ABC also provides consistent improvements over operational debiasing protocols, quadrupling the skill of the debiased SubX multimodel mean for precipitation (\textbf{a}) and doubling that of temperature (\textbf{b}).
        The average temporal skill over all forecast dates is displayed above each map.
        } \label{fig:skill_spatial_distribution_subx_mean}
        \end{figure}

        \begin{figure}[h!]
        \centering
        \includegraphics[width=\textwidth]{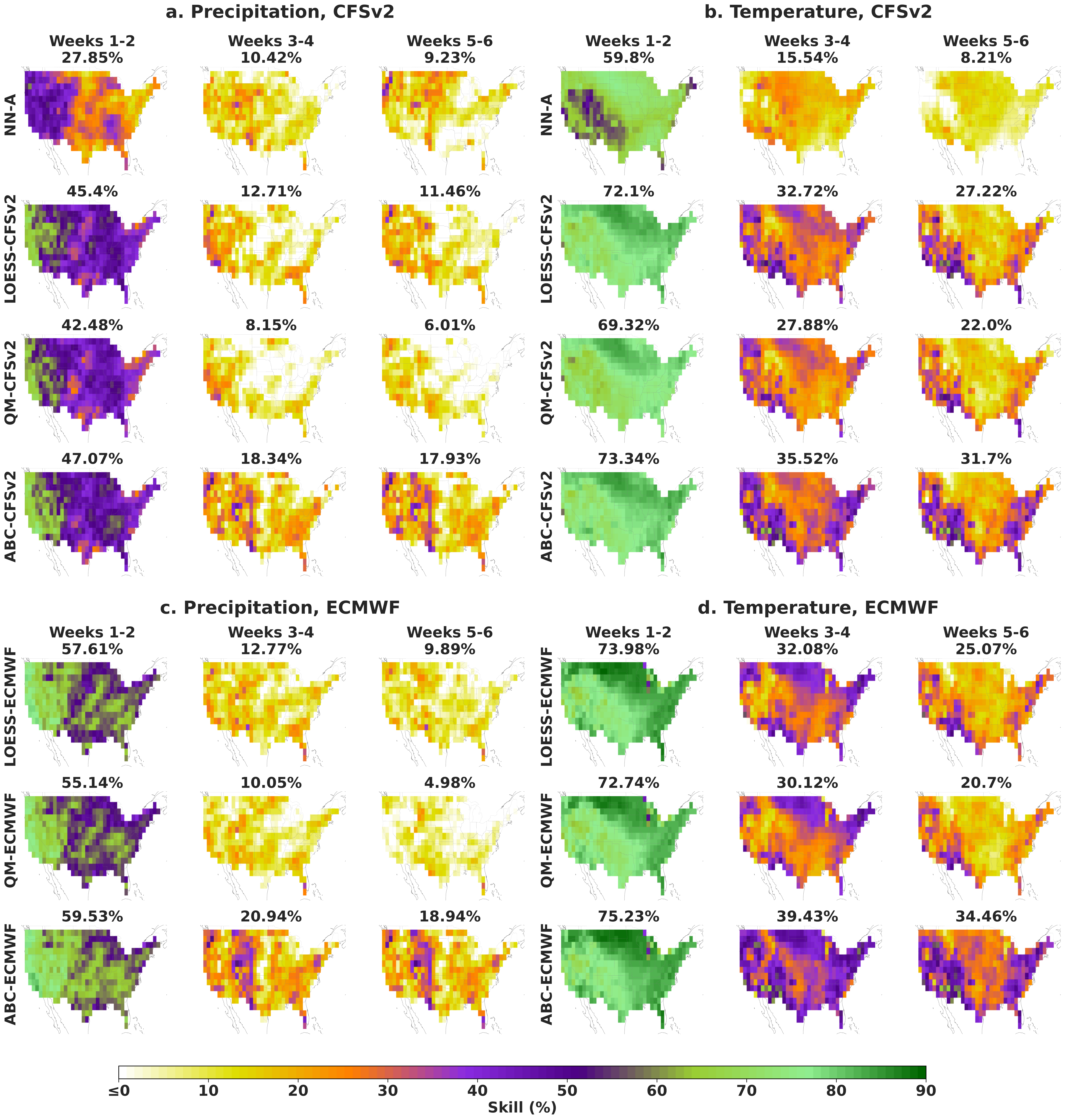}
        \caption{%
        \textbf{Spatial distribution of model skill for adaptive bias correction (ABC) and baseline neural network (NN-A), locally estimated scatterplot smoothing (LOESS), and quantile mapping (QM) corrections of dynamical models.} 
        For each grid point in the contiguous \US, spatial skill is averaged over the years 2018--2021.
        The average temporal skill over all forecast dates is displayed above each precipitation (\textbf{a, c}) and temperature (\textbf{b, d}) map.
        The NN-A correction operates specifically on \cfs model inputs.}
        \label{fig:skill_spatial_distribution_cfsv2}
\end{figure}

    \begin{figure}[th!]
        \centering
        \includegraphics[width=\textwidth]{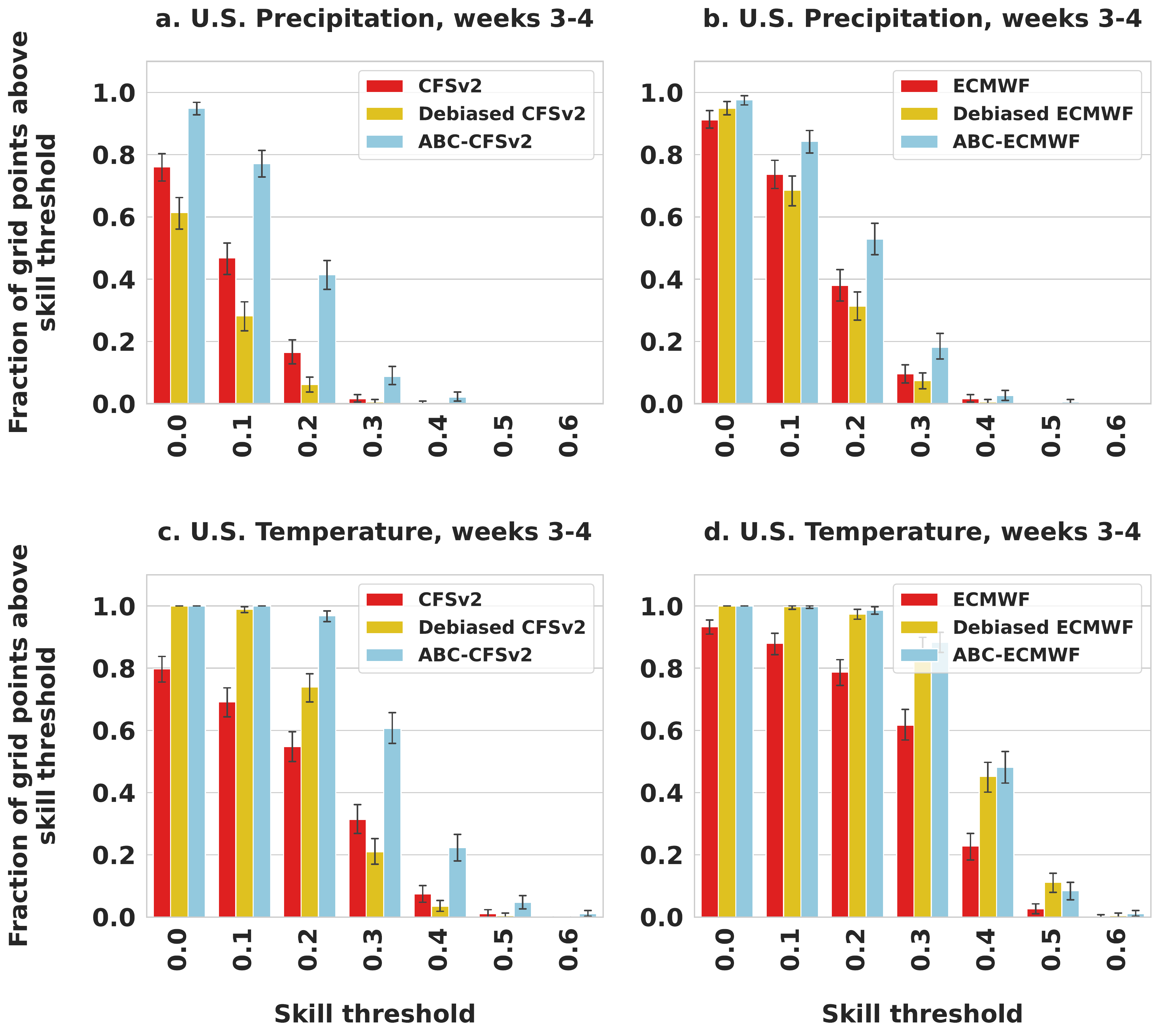}
    \caption{\textbf{Fraction of contiguous \US 
    with 2018--2021 spatial skill above a given threshold.}
    For each forecasting task and dynamical model input (\textbf{a, b, c, d}), 
    adaptive bias correction (ABC) consistently 
    expands the geographic range of 
    higher skill 
    over raw and operationally-debiased dynamical models.
    The error bars display 95\% bootstrap confidence intervals.%
    }    \label{fig:barplot_fraction_above_treshold_skill_34w}
    \end{figure}    

    \begin{figure}[th!]
        \centering
        \includegraphics[width=\textwidth]{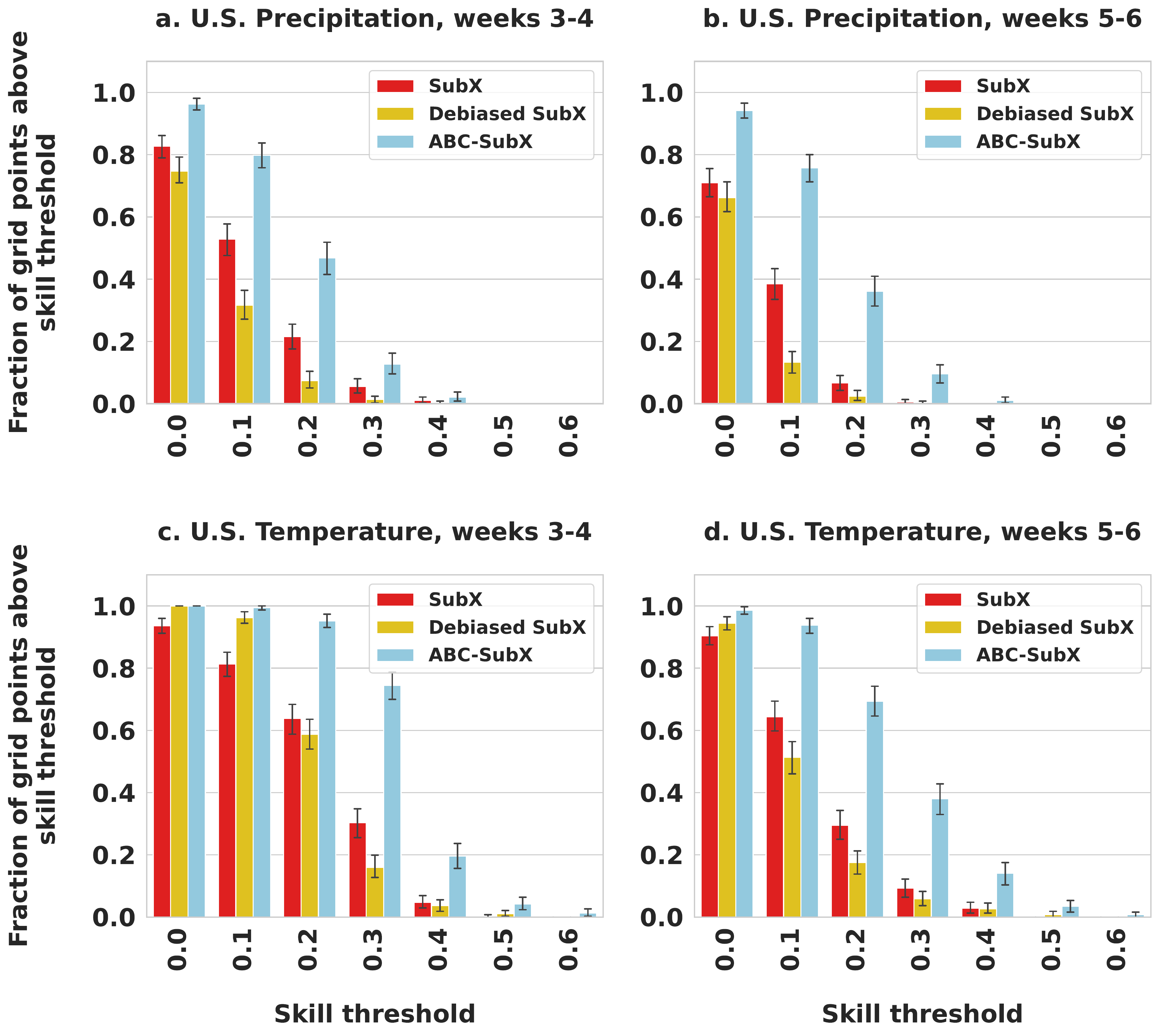}
    \caption{\textbf{Fraction of contiguous \US with 2018--2021 spatial skill above a given threshold.} 
    For each forecasting task (\textbf{a, b, c, d}), 
    adaptive bias correction (ABC) consistently 
    expands the geographic range of 
    higher skill over the raw and operationally-debiased SubX multimodel mean. 
    The error bars display 95\% bootstrap confidence intervals. 
    }    \label{fig:barplot_fraction_above_treshold_skill_subx_mean}
    \end{figure}

        \begin{figure}[h!]
        \centering
        \includegraphics[width=\textwidth]{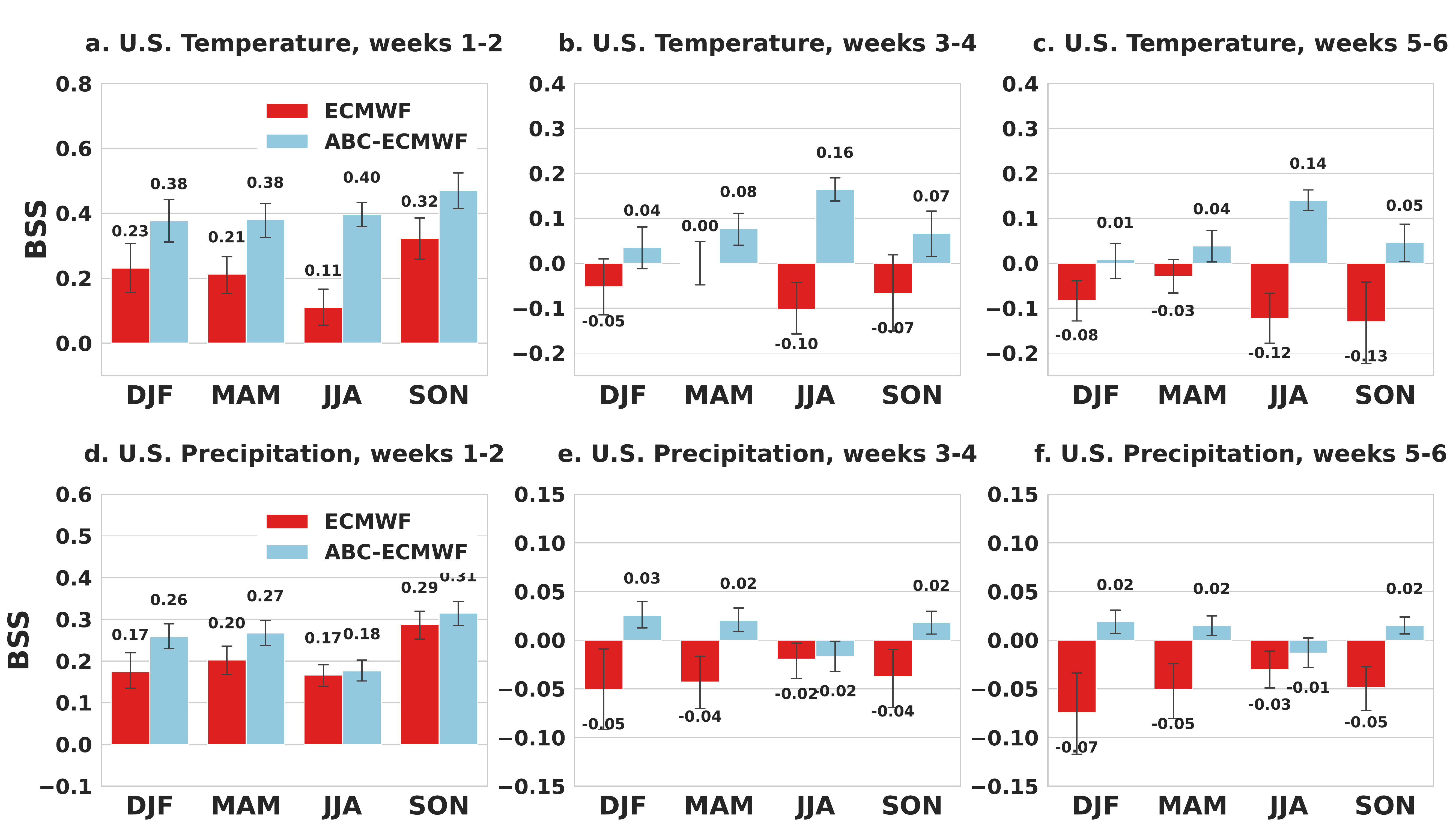}
        \caption{\textbf{Average Brier skill score (BSS) for above normal temperature or precipitation per season for \ecmwf (red) and its adaptive bias correction (ABC) counterpart (blue).}
        Higher BSS indicates a more skillful probabilistic forecast. 
        For each forecasting task (\textbf{a, b, c, d, e, f}), 
        per-date BSS is computed across the contiguous \US and averaged over the years 2018--2021. 
        Here, DJF = December, January, February; 
        MAM = March, April, May;
        JJA = June, July, August; and 
        SON = September, October, November.
        The error bars display 95\% bootstrap confidence intervals. 
        }
        \label{fig:barplot_bss_quaterly}
    \end{figure}

        \begin{figure}[h!]
        \centering
        \includegraphics[width=\textwidth]{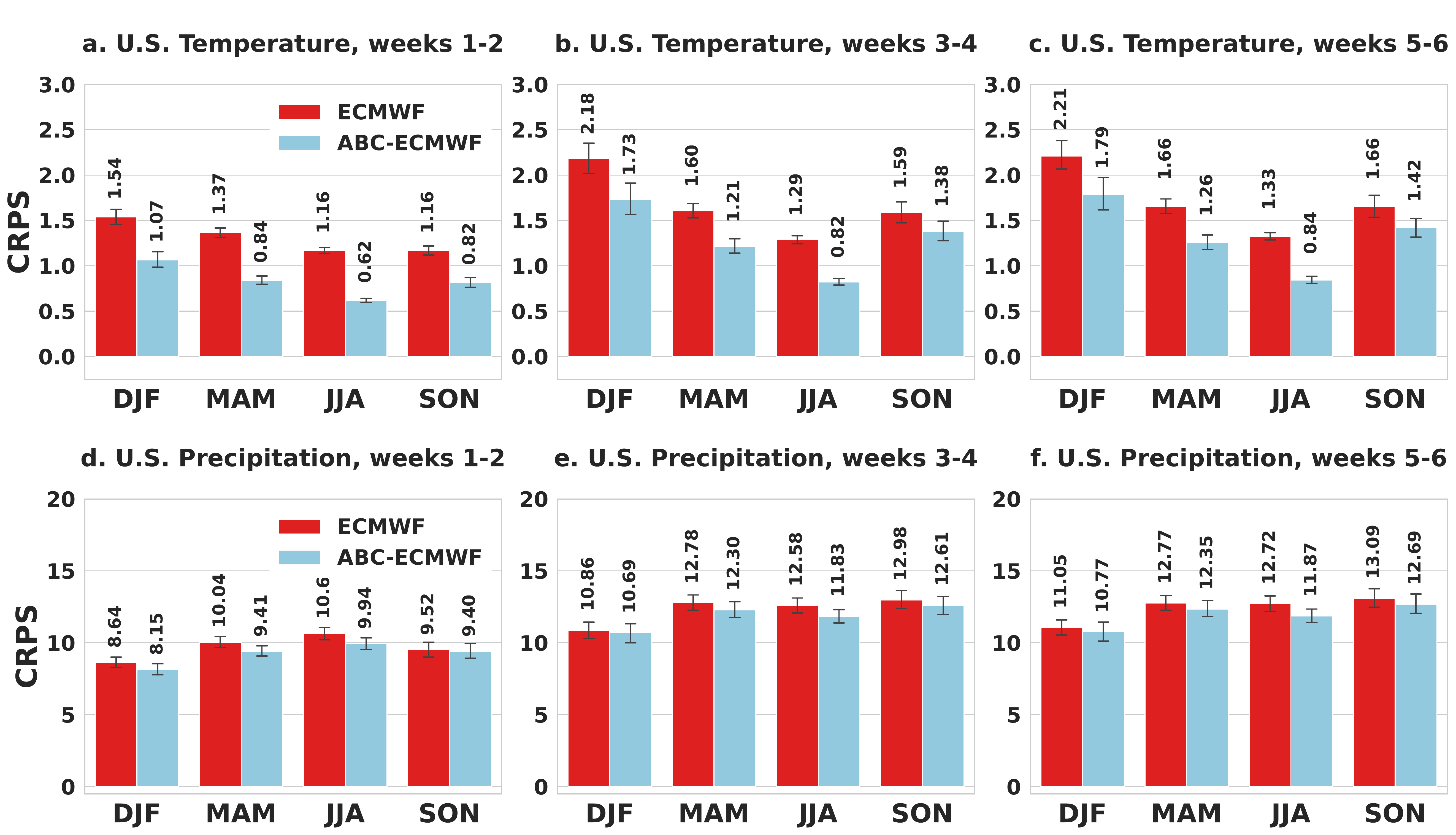}
        \caption{\textbf{Average continuous ranked probability score (CRPS) per season for \ecmwf (red) and its adaptive bias correction (ABC) counterpart (blue).}
        Lower CRPS indicates a more accurate probabilistic forecast. 
        For each forecasting task (\textbf{a, b, c, d, e, f}), 
        per-date CRPS is computed across the contiguous \US and averaged over the years 2018--2021. 
        Here, DJF = December, January, February; 
        MAM = March, April, May;
        JJA = June, July, August; and 
        SON = September, October, November.
        The error bars display 95\% bootstrap confidence intervals.}
        \label{fig:barplot_crps_quaterly}
    \end{figure}

       \begin{figure}[h!]
        \centering
        \includegraphics[width=\textwidth]{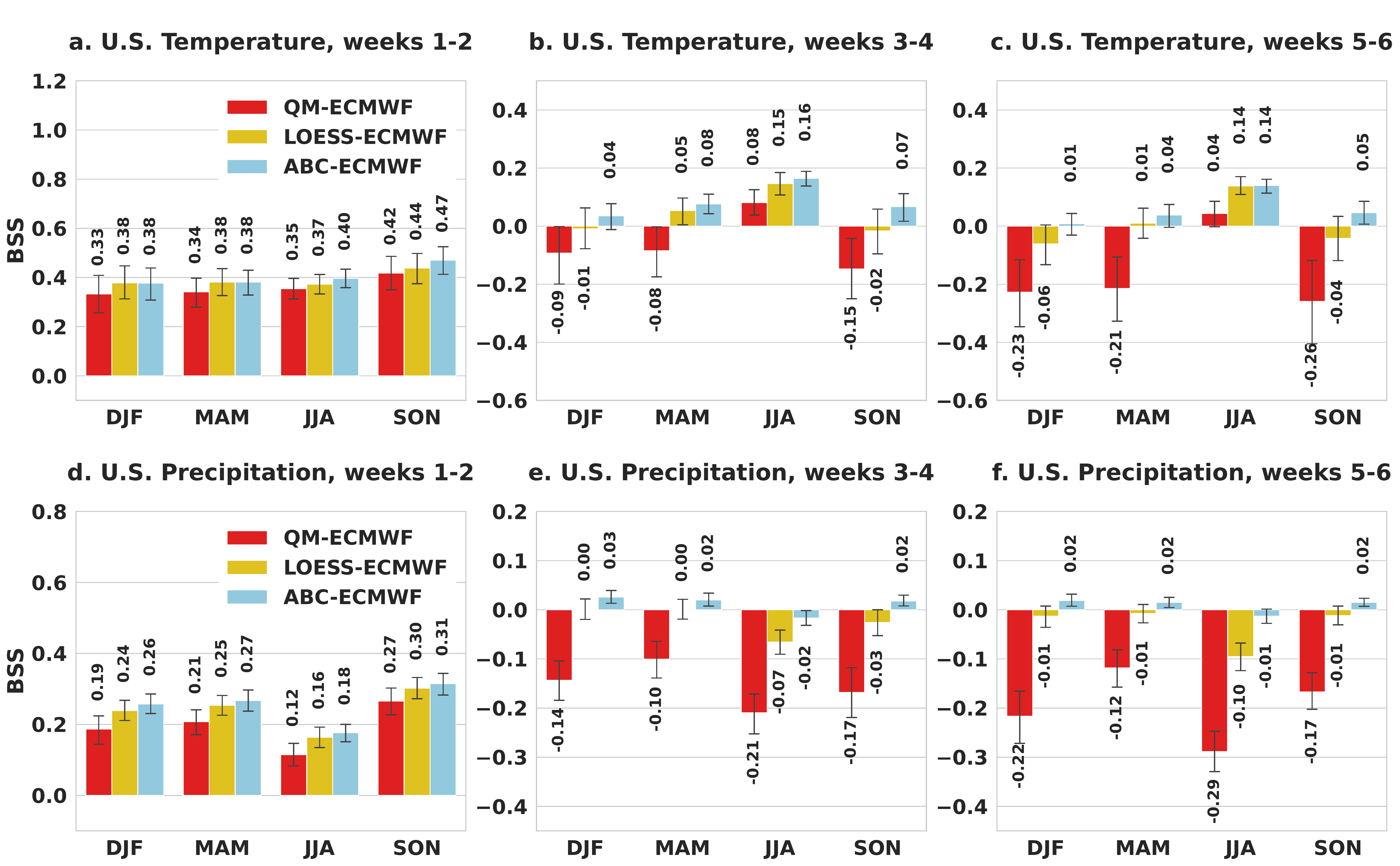}
        \caption{\textbf{Average Brier skill score (BSS) for above normal temperature or precipitation per season for adaptive bias correction (ABC) and baseline locally estimated scatterplot smoothing (LOESS) and quantile mapping (QM)  corrections of ECMWF.} 
        Higher BSS indicates a more skillful probabilistic forecast. 
        For each forecasting task (\textbf{a, b, c, d, e, f}), 
        per-date BSS is computed across the contiguous \US and averaged over the years 2018--2021. 
        Here, DJF = December, January, February; 
        MAM = March, April, May;
        JJA = June, July, August; and 
        SON = September, October, November.
        The error bars display 95\% bootstrap confidence intervals.}
        \label{fig:barplot_bss_quaterly_loess}
    \end{figure}

        \begin{figure}[h!]
        \centering
        \includegraphics[width=\textwidth]{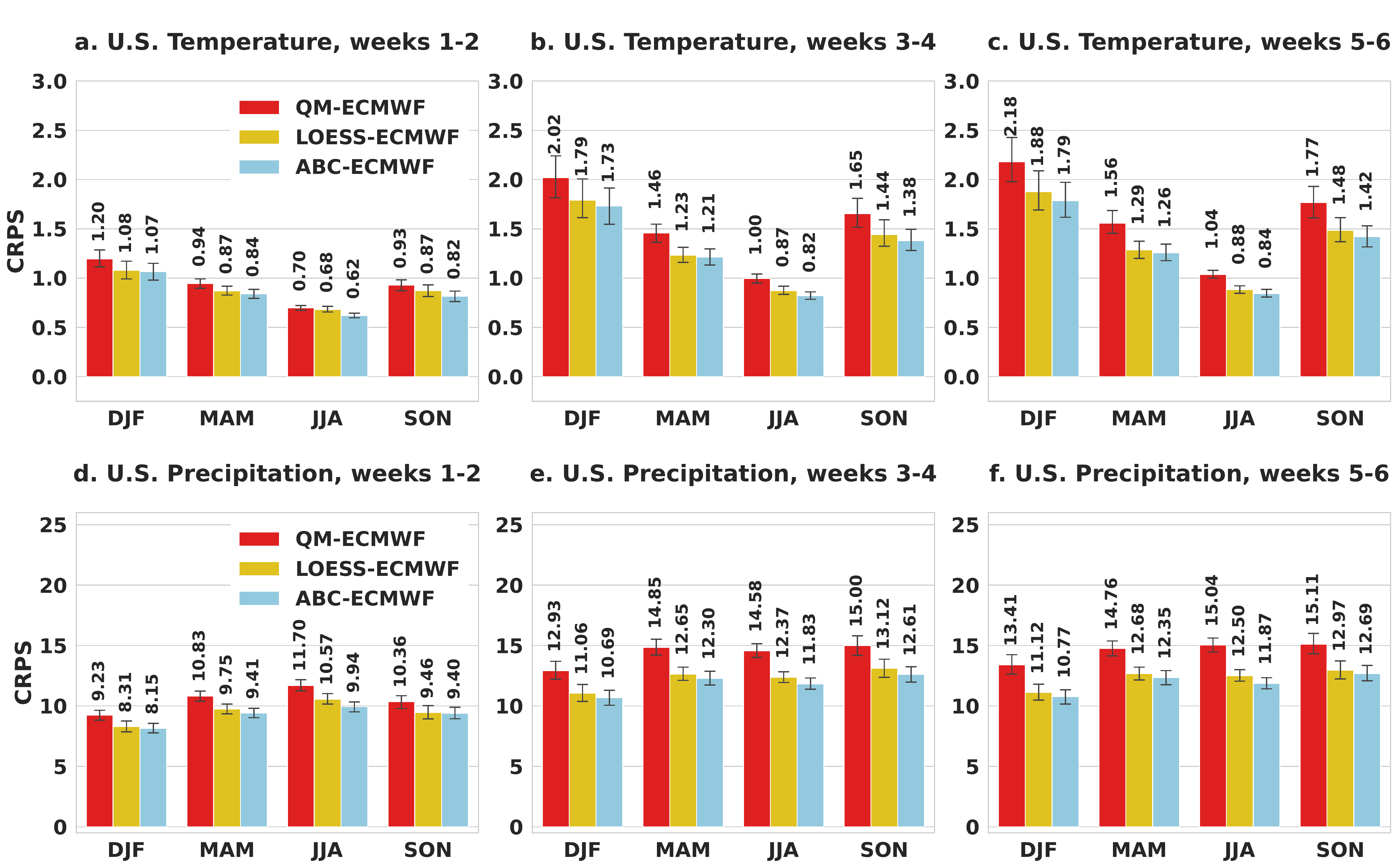}
        \caption{\textbf{Average continuous ranked probability score (CRPS) per season for 
        adaptive bias correction (ABC) and baseline locally estimated scatterplot smoothing (LOESS) and quantile mapping (QM)  corrections of ECMWF.}
        Lower CRPS indicates a more accurate probabilistic forecast. 
        For each forecasting task (\textbf{a, b, c, d, e, f}), 
        per-date CRPS is computed across the contiguous \US and averaged over the years 2018--2021. 
        Here, DJF = December, January, February; 
        MAM = March, April, May;
        JJA = June, July, August; and 
        SON = September, October, November.
        The error bars display 95\% bootstrap confidence intervals.}
        \label{fig:barplot_crps_quaterly_loess}
    \end{figure}

        \begin{figure}[!ht]
        \centering
        \includegraphics[width=\textwidth]{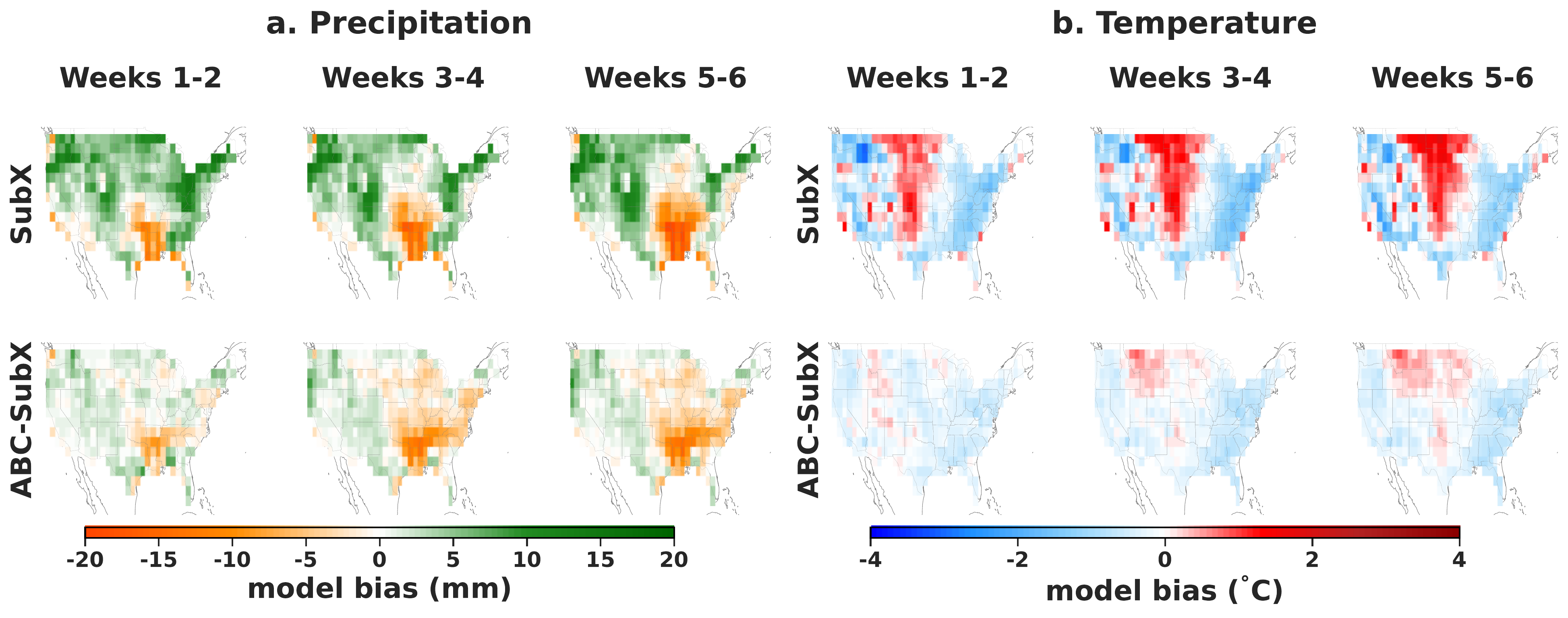}
        \caption{\textbf{Spatial distribution of model bias over the years 2018--2021.} Across the contiguous \US, adaptive bias correction (ABC) reduces the systematic model bias of the SubX multimodel mean input for both precipitation (\textbf{a}) and temperature (\textbf{b}).
        }
        \label{fig:bias_spatial_distribution_subx_mean}
        \end{figure}

        \begin{figure}[!ht]
        \centering
        \includegraphics[width=\textwidth]{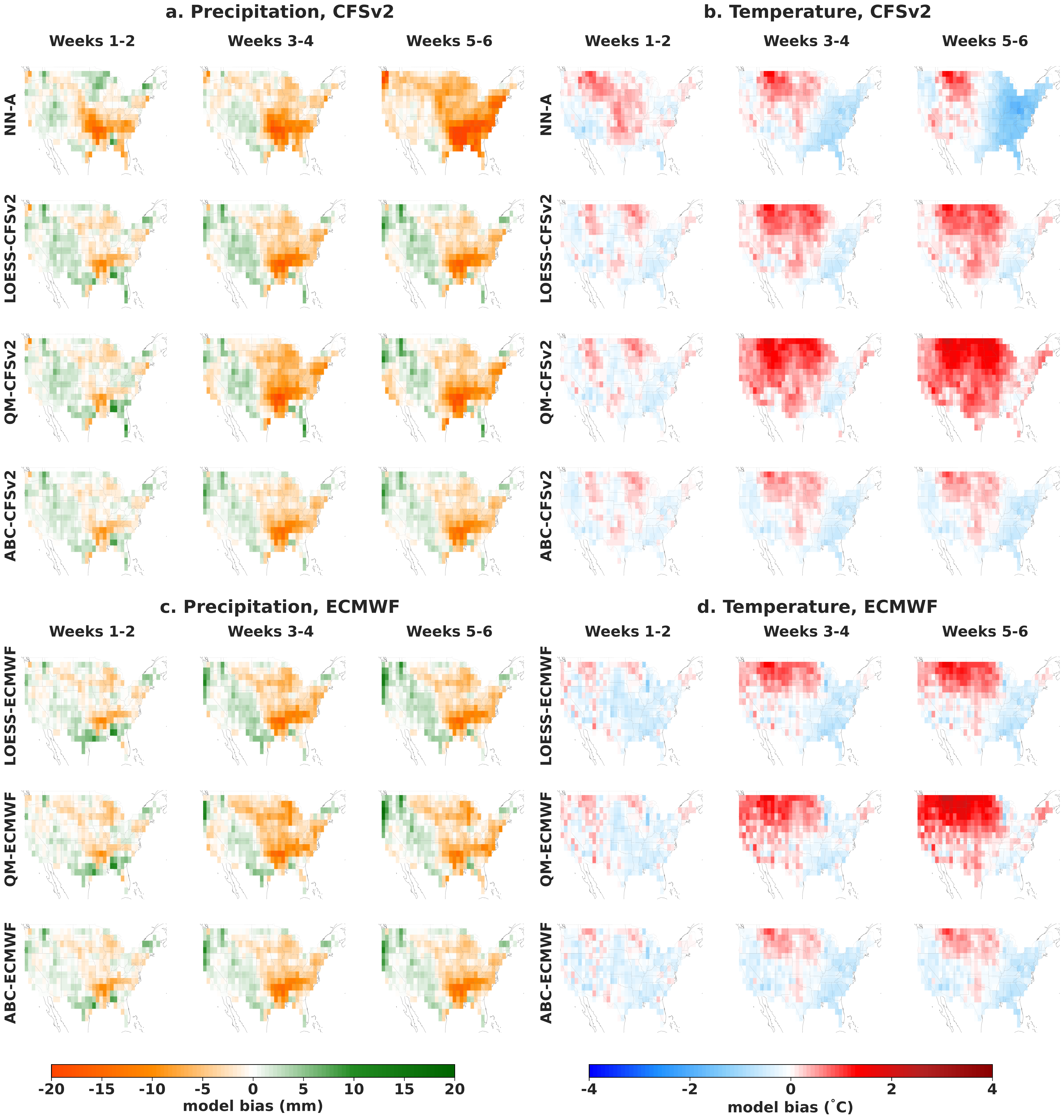}
        \caption{%
        \textbf{Spatial distribution of model bias for adaptive bias correction (ABC) and baseline neural network (NN-A), locally estimated scatterplot smoothing (LOESS) and quantile mapping (QM) corrections of dynamical models.} 
        For each grid point in the contiguous \US, bias is computed over years 2018--2021 for both precipitation (\textbf{a}, \textbf{c}) and temperature (\textbf{b}, \textbf{d}).
        The NN-A correction operates specifically on \cfs model inputs.}
        \label{fig:bias_spatial_distribution_cfsv2}
        \end{figure}

        \begin{figure}[!ht]
        \centering
        \includegraphics[width=\textwidth]{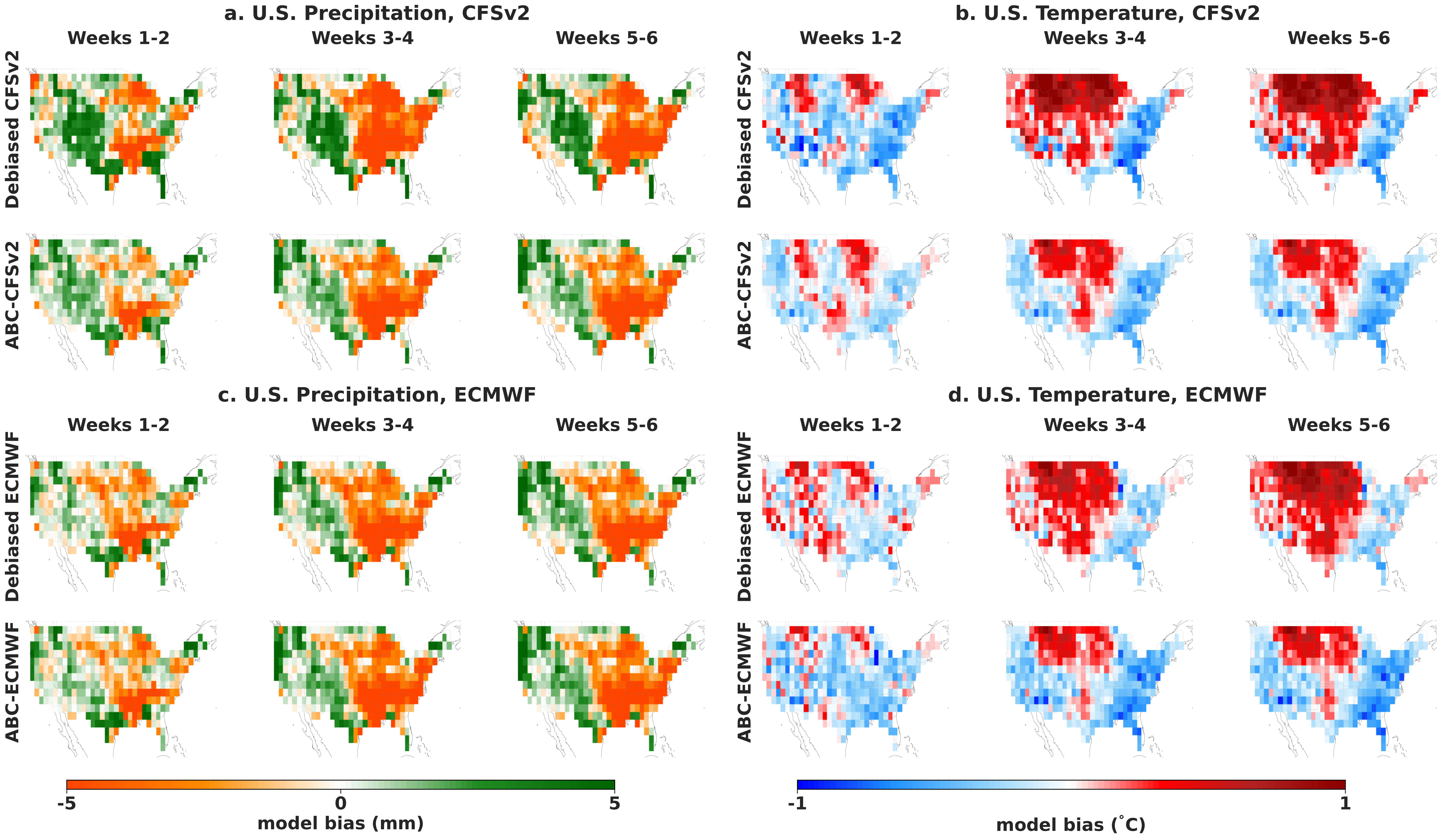}
        \caption{%
        \textbf{Spatial distribution of model bias for adaptive bias correction (ABC) and operationally-debiased dynamical models.} 
        For each grid point in the contiguous \US, bias is computed over years 2018--2021 for both precipitation (\textbf{a}, \textbf{c}) and temperature (\textbf{b}, \textbf{d}).
        }
        \label{fig:bias_spatial_distribution_deb_abc}
        \end{figure}

    \begin{figure}[h]
    \centering
    \includegraphics[width=0.98\textwidth]{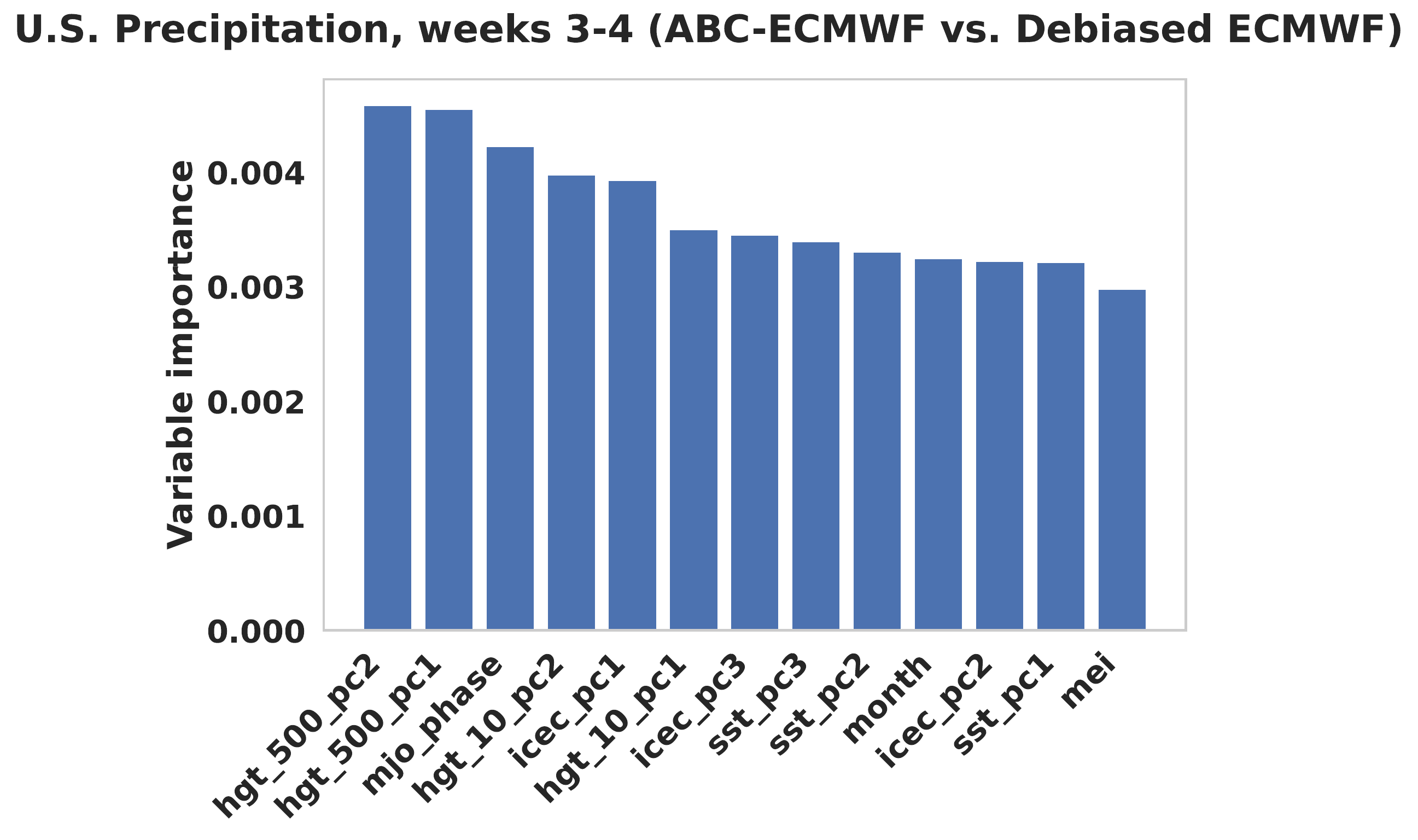}
    \caption{\textbf{Overall importance of explanatory variables in opportunistic adaptive bias correction (ABC) workflow.} The importance of each explanatory variable in explaining the weeks 3-4 precipitation skill improvement of \aecmwf over debiased \ecmwf is measured by Shapley effects.
    The explanatory variables considered are 
    the first two principal components (PCs) of 500 hPa geopotential height (\texttt{hgt\_500\_pc1} and \texttt{hgt\_500\_pc2}), 
    the first and second PC of 10 hPa geopotential height (\texttt{hgt\_10\_pc1} and \texttt{hgt\_10\_pc2}), 
    the first three PCs of sea ice concentration (\texttt{icec\_pc1}, \texttt{icec\_pc2} and \texttt{icec\_pc3}) and sea surface temperature (\texttt{sst\_pc1}, \texttt{sst\_pc2} and \texttt{sst\_pc3}), 
    the MJO phase (\texttt{mjo\_phase}), 
    the multivariate ENSO index (\texttt{mei}) 
    and the month (\texttt{month}). 
    }
    \label{fig:shapley_effects}
    \end{figure}

        \begin{figure}[h!]
        \centering
        \includegraphics[width=\textwidth]{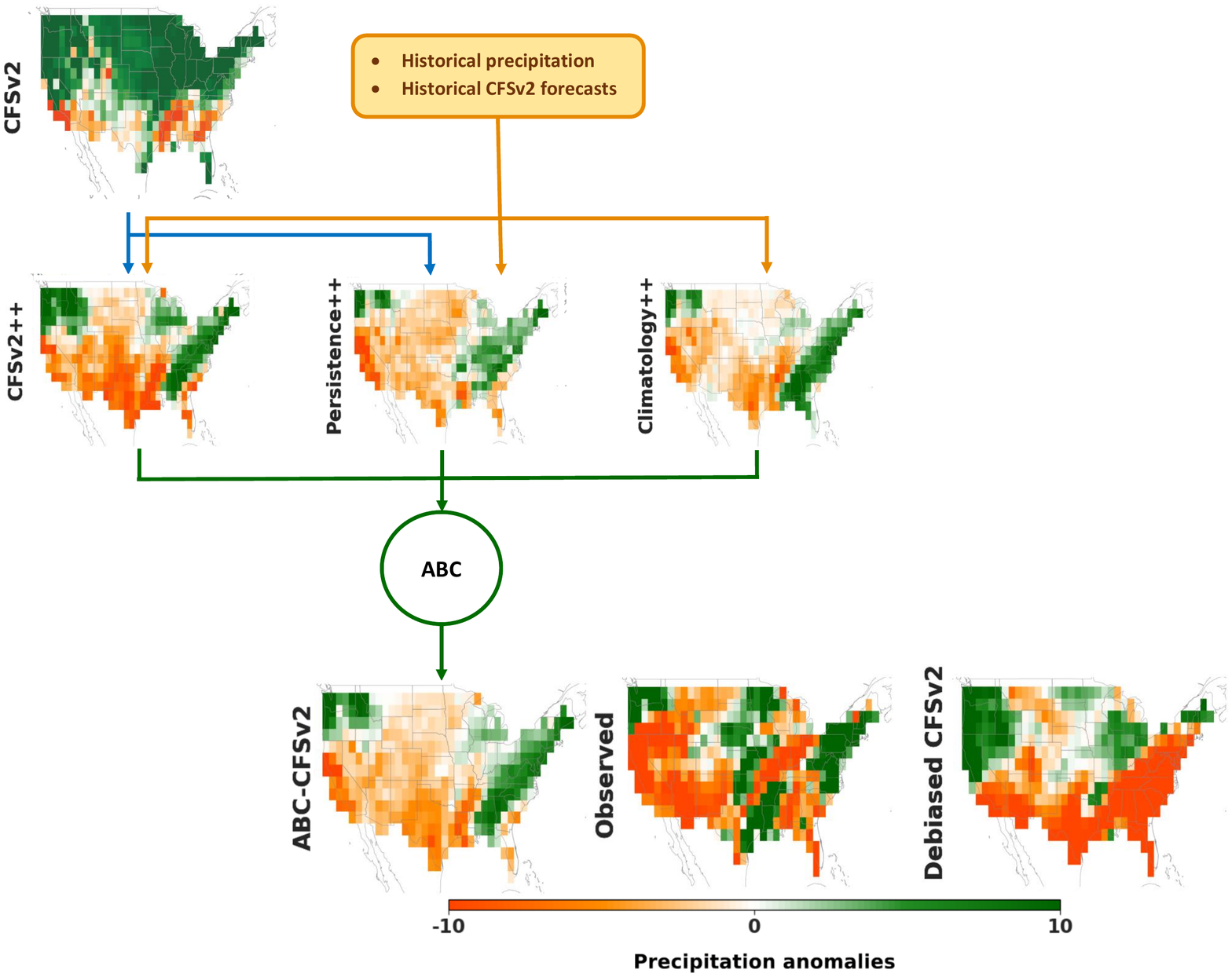}
        \caption{\textbf{Schematic of adaptive bias correction (ABC) model input and output data for weeks 3-4 precipitation forecasting.} Here, we compare ABC-corrected CFSv2 forecast with operationally debiased CFSv2 and the ground-truth observations for the target date 2020-12-18.}
        \label{fig:abc_schematic}
        \end{figure}
\renewcommand{\thealgorithm}{S\arabic{algorithm}}
\newpage
\clearpage
\section{Supplementary Methods}
\label{sec:supp_methods}
\subsection{ABC algorithm details}\label{sec:algorithms_details}
    This section presents the algorithm details for the three machine learning models underlying ABC:  \nwppp, \climpp, and \perpp.

   \begin{algorithm}%
  \caption{\nwppp}
  \label{alg:nwppp}
  \begin{algorithmic}
  
  	\INPUT test date $\tstar$; 
  	lead time $\lstar$; 
  	\# issuance dates $\dstar$;
    span $s$; 
    training set ground truth and \nwp forecasts $(\gt_t, \fcst_{t,l})_{t\in \trainset, l \in\leads}$
    \INIT days per year $D = 365.242199$; \# training years $Y = 12$
    \STATE $\mc{S} 
	= \{t \in \trainset:
	\texttt{year\_diff}\coloneqq \floor{\frac{t^\star-t}{D}} \leq Y
	\text{ and } 
	\texttt{day\_diff}\coloneqq \frac{365}{2} -|\floor{(t^\star-t)\mod D}- \frac{365}{2} | \leq s 
	\}$\\[1pt]
    \STATE // Form \nwp ensemble forecast across issuance dates and lead times $l\in\leads$
    \FOR{training and test dates $t \in \mc{S}\cup \{\tstar\}$}
    \STATE
    $
    \bar{\fcst}_t
    = \textup{mean}((\fcst_{t- \lstar - d + 1, l})_{1 \leq d \leq \dstar, l\in \leads})
    $ \\[2pt]
    \ENDFOR 
    \OUTPUT $\bar{\fcst}_{\tstar} + \textup{mean}((\gt_t - \bar{\fcst}_{t})_{t\in \mc{S}})$
  \end{algorithmic}
\end{algorithm}

    \begin{algorithm}%
  \caption{\climpp}
  \label{alg:climpp}
  \begin{algorithmic}
    \INPUT 
    test date $\tstar$;
    \# train years $Y$;
    span $s$;
    $\loss\!\in\!\{${\footnotesize$\rmse,\!\mse$}$\}$;
    training set ground truth $(\gt_t)_{t\in \trainset}$
    \\
    \INIT days per year $D = 365.242199$
    \STATE $\mc{S} 
	= \{t \in \trainset:
	\texttt{year\_diff}\coloneqq \floor{\frac{t^\star-t}{D}} \leq Y
	\text{ and } 
	\texttt{day\_diff}\coloneqq \frac{365}{2} -|\floor{(t^\star-t)\mod D}- \frac{365}{2} | \leq s 
	\}$
    \OUTPUT 
    $\argmin_{\gt} 
        \sum_{t \in \mc{S}} \loss(\gt, \gt_t)
    $
  \end{algorithmic}
\end{algorithm}
    \begin{algorithm}%
  \caption{\perpp}
  \label{alg:perpp}
  \begin{algorithmic}
  	\INPUT 
  	lead time $\lstar$; 
    training set ground truth, climatology, and \nwp forecasts $(\gt_t, \climvec_t, \fcst_{t,l})_{t\in \trainset, l \in\leads}$
    \\
    \INIT forecast period length $L = 14$\\
    \STATE // Form \nwp ensemble forecast across subseasonal lead times $l \geq \lstar$
    \FOR{training dates $t \in \trainset$}
    \STATE
    $
    \bar{\fcst}_t
    = \textup{mean}((\fcst_{t, l})_{l\geq \lstar})
    $ \\[2pt]
    \ENDFOR 
    \STATE // Combine ensemble forecast, climatology, and lagged measurements
    \FOR{grid points $g = 1$ {\bfseries to} $G$}
    \STATE
    $
    \mbi{\hat{\beta}}_g \in \argmin_{\mbi{\beta}} 
    \sum_{t \in \trainset}
    (y_{t,g} - {\mbi{\beta}}^\top{[1,c_{t,g}, y_{t-\lstar-L-1,g}, y_{t-2\lstar-L-1, g}, \bar{f}_{t-\lstar-1,g}]})^2
    $ \\[2pt]
    \ENDFOR
    \OUTPUT coefficients $(\mbi{\hat{\beta}}_g)_{g=1}^G$
  \end{algorithmic}
\end{algorithm}

\subsection{Probabilistic evaluation}
\label{sec:prob-skill}

\subsection*{ABC probabilistic forecast}
To construct a probabilistic forecast from the deterministic outputs of ABC, we first apply the \nwppp, \climpp, and \perpp models to the \nwp model ensemble mean to generate deterministic forecasts as usual.  
We then bias correct the forecast of each dynamical model ensemble member by adding the \nwppp deterministic forecast and subtracting the raw \nwp model ensemble mean forecast.
We then repeat this process using \perpp for bias correction in place of \nwppp.
After these bias correction steps, any negative forecasted precipitation values are replaced with $0$.
Finally, we use the empirical distribution over \nwppp-corrected ensemble members, \perpp-corrected ensemble members, and \climpp as our probabilistic forecast distribution.
In the case of \ecmwf, there are $51$ raw ensemble member forecasts (consisting of a single control forecast and $50$ perturbed forecasts), so the associated ABC-\ecmwf probabilistic forecast is an empirical distribution over $103$ corrections.

\subsection*{Baseline probabilistic forecasts}
To construct a probabilistic forecast from the deterministic output of a baseline debiasing procedure (like LOESS or quantile mapping), we first apply the debiasing procedure to the \nwp model ensemble mean to generate deterministic forecasts as usual.  
We then bias correct the forecast of each dynamical model ensemble member by adding the baseline deterministic forecast and subtracting the raw \nwp model ensemble mean forecast.
After this bias correction step, any negative forecasted precipitation values are replaced with $0$.
Finally, we use the empirical distribution over the corrected ensemble members as our probabilistic forecast distribution.

\subsection*{Continuous ranked probability score} 

For each target date $t$ and grid point $g$, the continuous ranked probability score (CRPS)~\citepSM{wilks2011statisticalsm} is measured as
\begin{align}
\textup{CRPS}(\hat{F}_{t,g}, y_{t,g}) = \int_{-\infty}^\infty (\hat{F}_{t,g}(x) - \indic{y_{t,g} \leq x})^2 dx
\end{align}
where $y_{t,g}$ represents the ground-truth observation for the date and grid point and $\hat{F}_{t,g}$ represents the predicted cumulative distribution function for the date and grid point.
A smaller CRPS value indicates a higher quality probabilistic forecast.

\subsection*{Brier skill score} 

For each target date $t$, grid point $g$, and user-supplied threshold $x_{t,g}$, the Brier score (BS)~\citepSM{wilks2011statisticalsm} is measured as
\begin{align}
    \textup{BS}(\hat{F}_{t,g}, y_{t,g}; x_{t,g}) = (\hat{F}_{t,g}(x_{t,g}) - \indic{y_{t,g} \leq x_{t,g}})^2
\end{align}
where $y_{t,g}$ represents the ground-truth observation for the date and grid point and $\hat{F}_{t,g}$ represents the predicted cumulative distribution function for the date and grid point.
For the threshold $x_{t,g}$, we choose the second tercile of the climatological distribution of observations from the same grid point, day, and month in the years 1981--2010, so that the BS equivalently measures the error of estimating the probability of above-normal (versus near-normal or below-normal) observations.
For this choice of thresholds, the Brier skill score (BSS)~\citepSM{wilks2011statisticalsm}, which compares the the average BS of a forecast to the average BS of climatology across all grid points, takes the form
\begin{align}
\textup{BSS}(\mbi{\hat{F}}_t, \gt_t; \mbi{x}_t) = 1 - \frac{\textfrac{1}{G}\textsum_{g=1}^g\textup{BS}(\hat{F}_{t,g}, y_{t,g}; x_{t,g})}{\textfrac{1}{G}\textsum_{g=1}^g(\textfrac{2}{3} - \indic{y_{t,g} \leq x_{t,g}})^2}
\end{align}
for $\gt_t \in \R^G$ the vector of ground-truth observations, $\mbi{x}_t\in\R^G$ the vector of thresholds, and $\mbi{\hat{F}}_{t}$ the collection of predicted cumulative distribution functions for each grid point.
A larger BSS value indicates a higher quality probabilistic forecast.

\bibliographystyleSM{IEEEtran}
\bibliographySM{references}
}
\end{document}